\documentclass{article}

\usepackage[final, nonatbib]{neurips_2023}

\usepackage[utf8]{inputenc} 
\usepackage[T1]{fontenc}    
\usepackage{hyperref}       
\usepackage{url}            
\usepackage{booktabs}       
\usepackage{amsfonts}       
\usepackage{nicefrac}       
\usepackage{microtype}      
\usepackage{xcolor}         

\usepackage{wrapfig}
\usepackage{amsmath}
\usepackage{amsthm}
\usepackage{algorithm}
\usepackage{algorithmic}
\usepackage{xcolor}
\usepackage{amssymb,amsfonts}
\usepackage{bbm, bm}
\usepackage{stackengine}
\usepackage{makecell}
\usepackage{multirow}
\usepackage{subfigure}
\usepackage{adjustbox}
\usepackage{comment}
\usepackage{enumitem}

\theoremstyle{definition}

\newcommand{\ourmodel}{\textsc{ConPE}}
\newcommand{\df}{\textit{domain factor}}
\newcommand{\dfs}{\textit{domain factors}}

\newcommand{\StateSet}{S}
\newcommand{\ActionSet}{A}

\DeclareMathOperator*{\argmax}{argmax}

\DeclareMathOperator*{\expectation}{\mathbb{E}}
\newcounter{row}
\makeatletter
\@addtoreset{subfigure}{row}
\makeatother
\usepackage[style=numeric, sorting=none]{biblatex}
\title{
    Efficient Policy Adaptation with Contrastive Prompt Ensemble for Embodied Agents
}

\author{%
  Wonje Choi, Woo Kyung Kim, SeungHyun Kim, Honguk Woo\thanks{Honguk Woo is the corresponding author.} \\
  Department of Computer Science and Engineering\\
  Sungkyunkwan University\\
  \texttt{\{wjchoi1995, kwk2696, kimsh571, hwoo\}@skku.edu} \\
}

\begin{document}

\maketitle

\begin{abstract}
For embodied reinforcement learning (RL) agents interacting with the environment, it is desirable to have rapid policy adaptation to unseen visual observations, but achieving zero-shot adaptation capability is considered as a challenging problem in the RL context.
To address the problem, we present a novel contrastive prompt ensemble ($\ourmodel$) framework which utilizes a pretrained vision-language model and a set of visual prompts, thus enabling efficient policy learning and adaptation upon a wide range of environmental and physical changes encountered by embodied agents.
Specifically, we devise a guided-attention-based ensemble approach with multiple visual prompts on the vision-language model to construct robust state representations. Each prompt is contrastively learned in terms of an individual domain factor that significantly affects the agent's egocentric perception and observation. For a given task, the attention-based ensemble and policy are jointly learned so that the resulting state representations not only generalize to various domains but are also optimized for learning the task.
Through experiments, we show that $\ourmodel$ outperforms other state-of-the-art algorithms for several embodied agent tasks including navigation in AI2THOR, manipulation in egocentric-Metaworld, and autonomous driving in CARLA, while also improving the sample efficiency of policy learning and adaptation.
\end{abstract}

\section{Introduction}
In the literature of vision-based reinforcement learning (RL), with the advance of unsupervised techniques and large-scale pretrained models for computer vision, the decoupled structure, in which visual encoders are separately trained and used later for policy learning, has gained popularity~\cite{drl:atc,drl:curl,drl:mpr}. This decoupling demonstrates high efficiency in low data regimes with sparse reward signals, compared to end-to-end RL. 
In this regard, several works on adopting the decoupled structure to embodied agents interacting with the environment were introduced~\cite{drl:pbit,drl:pgp}, and specifically, pretrained vision models (e.g., ResNet in~\cite{emb:zsogvn}) or vision-language models (e.g., CLIP in~\cite{emb:embclip,emb:zson}) were exploited for visual state representation encoders.
Yet, it is non-trivial to achieve zero-shot adaptation to visual domain changes in the environment with high diversity and non-stationarity, which are inherent for embodied agents. It was rarely investigated how to optimize those popular large-scale pretrained models to ensure the zero-shot capability of embodied agents. 

Embodied agents have several environmental and physical properties, such as egocentric camera position, stride length, and illumination, which are $\dfs$ making significant changes in agents' perception and observation. 
In the target (deployment) environment with uncalibrated settings on those domain factors, RL policies relying on pretrained visual encoders remain vulnerable to domain changes.

\begin{wrapfigure}{R}{.54\linewidth}
    \centering
    \includegraphics[width=.51\textwidth]{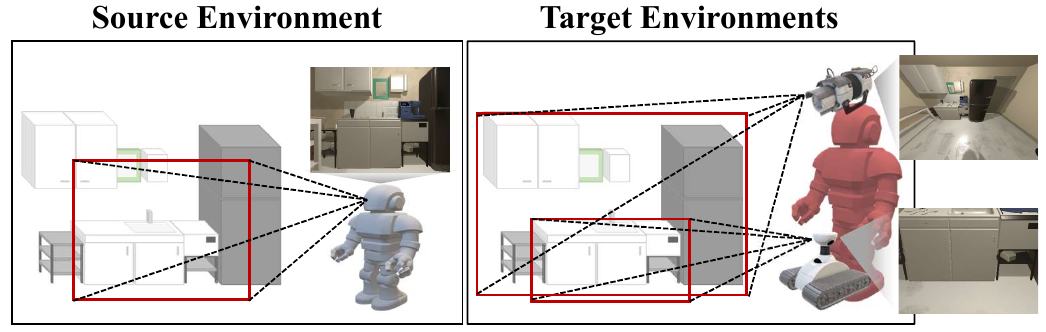}
    \caption{Visual Domain Changes of Embodied Agents}
    \label{fig:motivation}
\end{wrapfigure}

Figure~\ref{fig:motivation} provides an example of egocentric visual domain changes experienced by embodied agents due to different camera positions. When policies learned in the source environment are applied to the target environment, zero-shot performance can be significantly degraded, unless the visual encoder could adapt not only to environmental differences but also to the physical diversity of agents. 
In this paper, we investigate RL policy adaptation techniques for embodied agents to enable zero-shot adaptation to domain changes, by leveraging prompt-based learning for pretrained models in the decoupled RL structure. 
To this end, we present $\ourmodel$, a novel contrastive prompt ensemble framework that uses the CLIP vision-language model as the visual encoder, and facilitates dynamic adjustments of visual state representations against domain changes through an ensemble of contrastively learned visual prompts.
In $\ourmodel$, the ensemble employs attention-based state composition on multiple visual embeddings from the same input observation, where each embedding corresponds to a state representation individually prompted for
\color{black}
a specific domain factor. Specifically, the cosine similarity between an input observation and its respective prompted embeddings is used to calculate attention weights effectively.  

Through experiments, we demonstrate the benefits of our approach. 
First, RL policies learned via $\ourmodel$ achieve competitive zero-shot performance upon a wide variety of egocentric visual domain variations for several embodied agent tasks, such as  navigation tasks in AI2THOR~\cite{env:ai2thor}, vision-based robot manipulation tasks in egocentric-Metaworld, and autonomous driving tasks in CARLA~\cite{env:carla}. 
For instance, the policy via $\ourmodel$ outperforms EmbCLIP~\cite{emb:embclip} in zero-shot performance by 20.7\% for unseen target domains in the AI2THOR object navigation.
Second, our approach achieves high sample-efficiency in the decoupled RL structure. For instance, $\ourmodel$ requires less than 50.0\% and 16.7\% of the samples compared to ATC~\cite{drl:atc} and 60\% and 50\% of the samples compared to EmbCLIP to achieve comparable performance in seen and unseen target domains in the AI2THOR object navigation.

In the context of RL, our work is the first to explore policy adaptation using visual prompts for embodied agents, achieving superior zero-shot performance and high sample-efficiency.
The main contributions of our work are as follows. 
\begin{itemize}[]
\item We present a novel $\ourmodel$ framework with an ensemble of visual contrastive prompts, which enables zero-shot adaptation for vision-based embodied RL agents.
\item We devise visual prompt-based contrastive learning and guided-attention-based prompt ensemble algorithms to represent task-specific information in the CLIP embedding space.
%
\item We experimentally show that policies via $\ourmodel$ achieve comparable or superior zero-shot performance, compared to other state-of-the-art baselines, for several tasks.
We also demonstrate high sample-efficiency in policy learning and adaptation.
\item We create the datasets with various visual domains in AI2THOR, egocentric-Metaworld and CARLA, and make them publicly accessible for further research on RL policy adaptation.
\end{itemize}
\section{Problem Formulation}
In RL formulation, a learning environment is defined as a Markov decision process (MDP) of $(\StateSet,\ActionSet,\mathcal{P},R)$ with state space $s \in \StateSet$, action space $a \in \ActionSet$, transition probability $\mathcal{P}: \StateSet \times \ActionSet \rightarrow \StateSet$ and reward function $R: \StateSet \times \ActionSet \rightarrow \mathbb{R}$.  The objective of RL is to find an optimal policy $\pi^*: \StateSet \rightarrow \ActionSet$ maximizing the sum of discounted rewards.
For embodied agents, states might not be fully observable, and the environment is represented by a partially observable MDP (POMDP) of a tuple $(\StateSet, \ActionSet, \mathcal{P}, R, \Omega, \mathcal{O})$ with an observation space $o \in \Omega$ and a conditional observation probability~\cite{sutton2018reinforcement} $\mathcal{O}: \StateSet \times \ActionSet \rightarrow \Omega$. 

Given visual domains in the dynamic environment, we consider policy adaptation to find the optimal policy that remains invariant across the domains or is transferable to some target domain, where each domain is represented by a POMDP and domain changes are formulated by different $\mathcal{O}$. We denote domains as $D = (\Omega, \mathcal{O})$. 
Aiming to enable zero-shot adaptation to various domains, we formulate the policy adaptation problem as finding the optimal policy $\pi^*$ such that
\begin{equation}
    \pi^* = \argmax_\pi \left[ \expectation_{D \sim p(D)} \left[\sum_{t=1}^\infty\gamma^t R(s_t, \pi(o_t))\right]\right]
\end{equation}
where $p(D)$ is a given domain distribution and $\gamma$ is a discount factor of the environment.  

For embodied agents, the same state can be differently observed depending on the configuration of properties such as egocentric camera position, stride length, illumination, and object style.
We refer to such a property causing domain changes in the environment as a $\df$. 
Practical scenarios often involve the interplay of multiple domain factors in the environment.

\begin{figure}[t]
\centering
\includegraphics[width=0.99\linewidth]{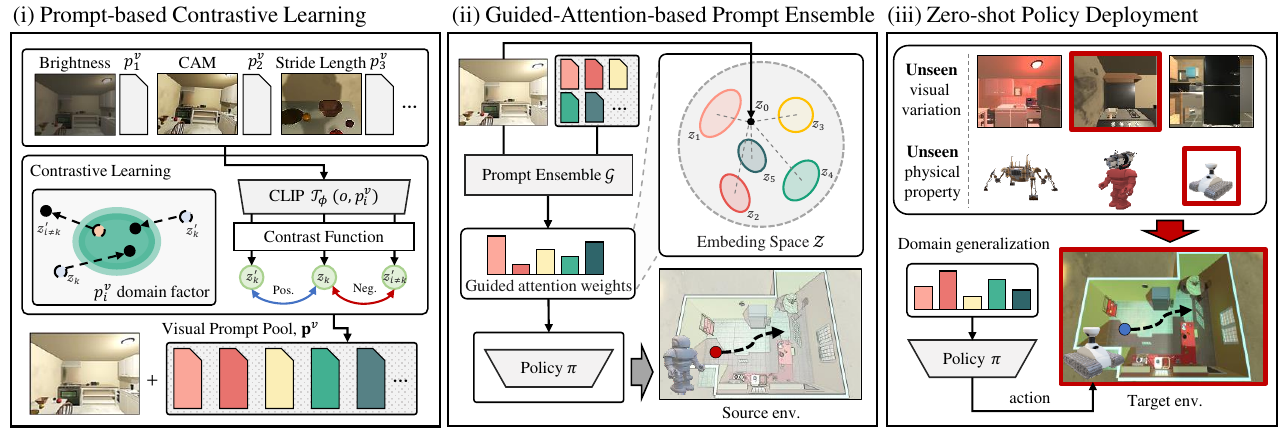}
\caption{$\ourmodel$ Framework. The  CLIP visual encoder is enhanced offline via (\romannumeral 1) prompt-based contrastive learning that generates the visual prompt pool, and a policy is learned online by (\romannumeral 2) guided-attention-based prompt ensemble that uses the prompt pool. In (\romannumeral 3) zero-shot deployment, the policy is immediately evaluated upon domain changes.}
\label{fig:Framework}
\end{figure}

\section{Our Approach}

\subsection{Framework Structure}
To enable zero-shot policy adaptation to unseen domains, we develop the $\ourmodel$ framework consisting of
(\romannumeral 1) prompt-based contrastive learning with the CLIP visual encoder,
(\romannumeral 2) guided-attention-based prompt ensemble, and
(\romannumeral 3) zero-shot policy deployment,
as illustrated in Figure~\ref{fig:Framework}.
The capability of the CLIP visual encoder is enhanced using multiple visual prompts that are contrastively learned on expert demonstrations for several domain factors. This establishes the visual prompt pool in (\romannumeral 1).
Then, the prompts are used to train the guided-attention-based ensemble with the environment in (\romannumeral 2).
To enhance learning efficiency and interpretability of attention weights, we use the cosine similarity of embeddings.
The attention module and policy are jointly learned for a specific task so that resulting state representations tend to generalize across various domains and be optimized for task learning.
In deployment, a non-stationary environment where its visual domain varies according to the environment conditions and agent physical properties is considered, and the zero-shot performance is evaluated in (\romannumeral 3).

\subsection{Prompt-based Contrastive Learning}\label{subsec:conpromptpoollearning}
To construct domain-invariant representations with respect to a specific domain factor for egocentric perception data, 
we adopt several contrastive tasks for visual prompt learning, which can be learned on a few expert demonstrations. For this, we use a visual prompt 
\begin{equation}
    p^v = [e^v_1, e^v_2,...,e^v_u],\ e^v_i \in \mathbb{R}^{d}
\end{equation}
where $e^v_i$ is a continuous learnable vector with the image patch embedding dimension $d$ (e.g., 768 for CLIP visual encoder) and $u$ is the length of a visual prompt.
Let a pretrained model $\mathcal{T}_{\phi}$ parameterized by $\phi$ maps observations $o \in \Omega$ to the embedding space $\mathcal{Z}$. 
With a contrast function $P: \Omega \times \Omega \rightarrow \{0, 1\}$~\cite{drl:atc, drl:curl, drl:aco} to discriminate whether an observation pair is positive or not, consider an $m$-sized batch of observation pairs $\mathcal{B}_P = \{(o_i, o_i')\}_{i\leq m}$ containing one positive pair $\{(o_k, o_k') | P(o_k, o_k') = 1\}$ for some $k \leq m$. 
Then, we enhance the capability of $\mathcal{T}_{\phi}$ by learning a visual prompt $p^v$ through contrastive learning, 
where the contrastive loss function~\cite{con:coding} is defined as
\begin{equation}
    \mathcal{L}_{\text{CON}}(p^v, \mathcal{B}_P) = -\log \left(\frac
    {S(\mathcal{T}_\phi(o_k, p^v), \mathcal{T}_\phi(o_k', p^v))} 
    {\sum_{i \neq k} S(\mathcal{T}_\phi(o_i, p^v), \mathcal{T}_\phi(o_i', p^v))}\right),\ 
    S(x, y) = \frac{1}{\lambda} \exp\left(\frac{\langle x, y \rangle}{\| x \| \| y \|}\right).
\label{loss:ContrastiveLoss}
\end{equation}
As in~\cite{clr}, for latent vectors $x,\ y \in \mathcal{Z}$, their similarity in the embedding space $\mathcal{Z}$ is calculated by $S(x, y)$, where $\lambda$ is a hyperparameter.
By conducting the prompt-based contrastive learning on $n$ different domain factors, we obtain a visual prompt pool 
\begin{equation}
    \mathbf{p}^v = [p^v_1, p^v_2, ... , p^v_n].
\end{equation}
Through this process, each visual prompt in $\mathbf{p}^v$ encapsulates domain-invariant knowledge pertinent to its respective domain factor.

\begin{figure}[t]
\centering
\includegraphics[width=0.92\linewidth]{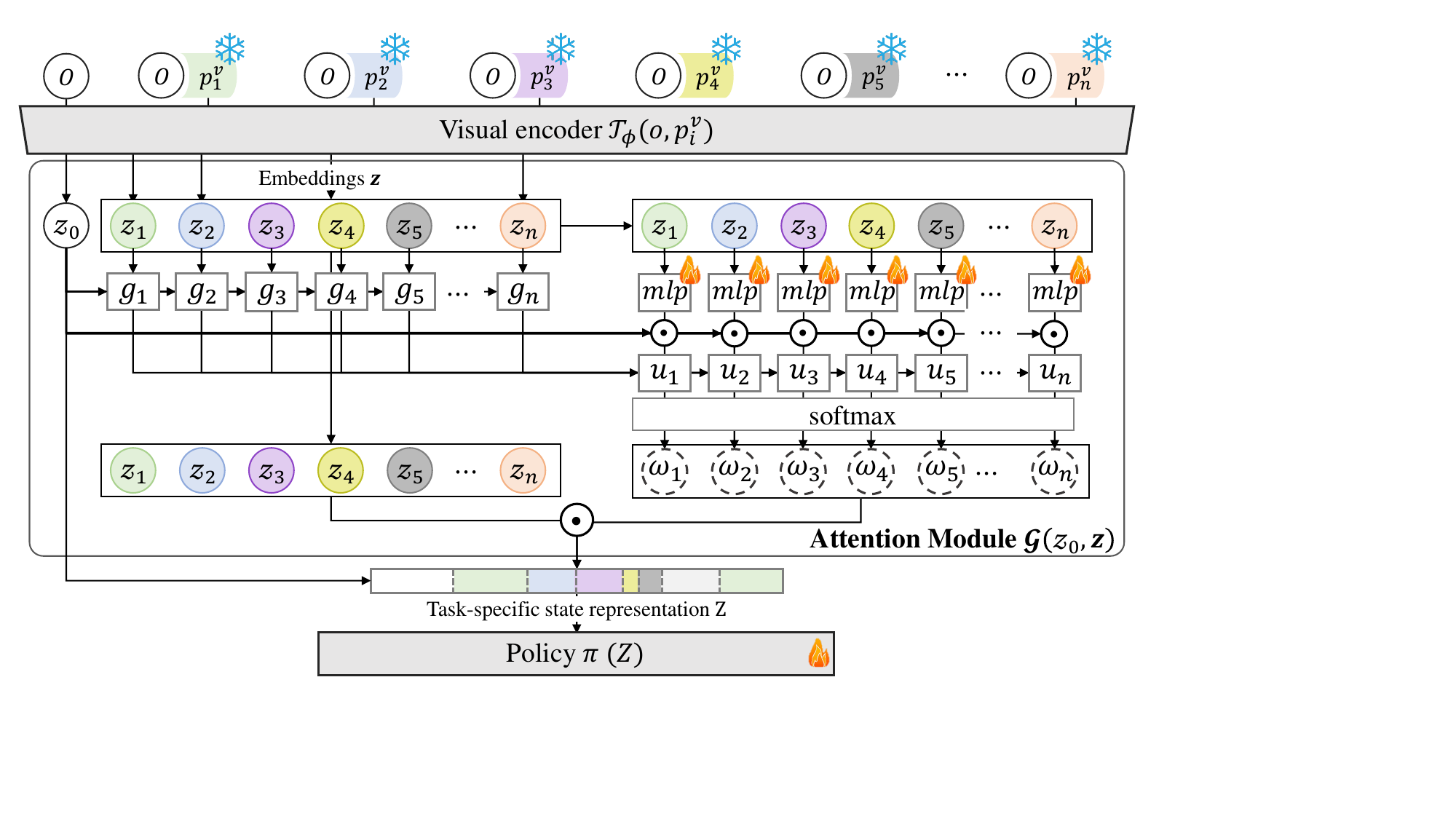}
\caption{Guided-Attention-based Prompt Ensemble. The cosine similarity-guided attention module $\mathcal{G}$ yields task-specific state representations from multiple prompted embeddings and is learned with a policy network $\pi$.} 
\label{fig:Method}
\end{figure}

\subsection{Guided-Attention-based Prompt Ensemble}\label{subsec:attention-based}
To effectively integrate individual prompted embeddings from multiple visual prompts into a task-specific state representation, we devise a guided-attention-based prompt ensemble structure, as shown in Figure~\ref{fig:Method} where the attention weights on the embeddings are dynamically computed via the attention module $\mathcal{G}$ for each observation. 

Given observation $o$ and the learned visual prompt pool $\mathbf{p}^v$, an image embedding $z_0 = \mathcal{T}_\phi(o)$ and prompted embeddings $\mathbf{z} = [z_1=\mathcal{T}_\phi(o,p^v_1),...,z_n]$ are calculated. 
Then, $z_o$ and $\mathbf{z}$ are fed to the attention module $\mathcal{G}$, where attention weights $\omega_i$ for each prompted embedding $z_i$ are optimized.
Since directly computing the attention weights using $z_0$ and $\mathbf{z}$ is prone to have an uninterpretable local optima, we introduce a guidance score $g_i$ based on the cosine similarity between the input image and visual prompted image embeddings in $\mathcal{Z}$, i.e., $g_i = \frac{\langle z_0, z_i \rangle}{\| z_0 \|\| z_i \|}$.
Given that larger $g_i$ signifies a stronger conformity of an observation to the domain factor relevant to the prompted embedding $z_i$, we use $g_i$ to steer the attention weights, aiming to not only improve learning efficiency but also provide interpretability.
With guidance $g_i$, we compute the attention weights $\omega_i$ by
\begin{equation} \label{equ:gattn}
    \omega_{i} = \frac{\text{exp}(u_i/\tau)}{\sum_{k}\text{exp}(u_k/\tau)},  \ \ \ 
    u_i =  \frac{\langle z_0, k_i \rangle }{\sqrt{d}}  g_i
\end{equation}
where $k_i$ is the projection of $z_i$, $d$ is dimension of $z$, and $\tau$ is a softmax temperature. 
Then, state embedding $Z$ is obtained by
\begin{equation} \label{equ:embZ}
    Z = \mathcal{G}(z_0, \mathbf{z}) = z_0 + \sum_{i=1}^n \omega_i z_i.
\end{equation}

Algorithm~\ref{alg:framework} shows the procedures in $\ourmodel$, where the first half corresponds to prompt-based contrastive learning (in Section~\ref{subsec:conpromptpoollearning}) and the other half corresponds to joint learning of a policy $\pi(Z)$ and the attention module $\mathcal{G}$. 
As $\mathcal{G}$ is optimized by a given RL task objective in the source domains (in line 12), the resulting $Z$ tends to be task-specific, while $Z$ is also domain-invariant by the ensemble of contrastively learned visual prompts based on $\mathcal{G}$ with respect to the combinations of multiple domain factors. 
The entire algorithm can be found in Appendix.
\begin{algorithm}[t]
\caption{Procedure of $\ourmodel$ Framework}
\label{alg:framework}
Dataset $\mathcal{D}= \{(o_1, o'_1), ...\}$, replay buffer $Z_D \leftarrow \emptyset$, pretrained vision-language model $\mathcal{T}_\phi$ \\
Visual prompt pool $\mathbf{p}^v = [p^v_1,...,p^v_n]$, attention module $\mathcal{G}$, policy $\pi$
\begin{algorithmic}[1]
\STATE \textit{/* Prompt-based Contrastive Learning */}
\FOR{$i = 1, ..., n$}
    \WHILE{not converge}
        \STATE{Sample a batch $\mathcal{B}_{P_i} = \{(o_{j},o_{j}')\}_{j\leq m} \sim \mathcal{D}$}
        \STATE{Update prompt $p^v_i \gets p^v_i - \nabla\mathcal{L}_{\text{CON}}(p^v_i, \mathcal{B}_{P_i})$ using~\eqref{loss:ContrastiveLoss}}
    \ENDWHILE
\ENDFOR
\STATE \textit{/* Prompt Ensemble-based Policy Learning */} \\
\FOR{each environment step}
    \STATE{Sample action $a = \pi(\mathcal{G}(\mathcal{T}_{\phi}(o), \textbf{z}))$ using~\eqref{equ:gattn},~\eqref{equ:embZ}}
    \STATE{$Z_D \gets Z_D \cup \{(\textbf{z}, a, r)\}$}
    \STATE{Jointly optimize policy $\pi$ and module $\mathcal{G}$ on $\{(\textbf{z}_j, a_j, r_j)\}_{j\leq m}\sim Z_{D}$}
\ENDFOR
\end{algorithmic}
\end{algorithm}

\section{Evaluation}
\noindent\textbf{Experiments.} 
We use AI2THOR~\cite{env:ai2thor}, Metaworld~\cite{env:metaworld}, and CARLA ~\cite{env:carla} environments, specifically configured for embodied agent tasks with dynamic domain changes. 
These environments allow us to explore various $\dfs$ such as camera settings, stride length, rotation degree, gravity, illuminations, wind speeds, and others.
For prompt-based contrastive learning (in Section~\ref{subsec:conpromptpoollearning}), we use a small dataset of expert demonstrations for each domain factor (i.e., 10 episodes per domain factor).
For prompt ensemble-based policy learning (in Section~\ref{subsec:attention-based}), we use a few source domains randomly generated through combinatorial variations of the seen domain factors (i.e., 4 source domains).
In our zero-shot evaluations, we use target domains that can be categorized as either seen or unseen. The seen target domains are those encountered during the phase of prompt-based contrastive learning, while these domains are not present during the phase of prompt ensemble-based policy learning.
On the other hand, the unseen target domains refer to those that are entirely new, implying that they are not encountered during either learning phases.

\noindent\textbf{Baselines.} 
We implement several baselines for comparison.
LUSR~\cite{drl:lusr} is a reconstruction-based domain adaptation method in RL, which uses the variational autoencoder structure for robust representations.
CURL~\cite{drl:curl} and ACT~\cite{drl:atc} employ contrastive learning in RL frameworks for high sample-efficiency and generalization to visual domains.
ACO~\cite{drl:aco} utilizes augmentation-driven and behavior-driven contrastive tasks in the context of RL. 
EmbCLIP~\cite{emb:embclip} is a state-of-the-art embodied AI model, which exploits the pretrained CLIP visual encoder for visual state representations. 

\noindent\textbf{Implementation.} 
We implement $\ourmodel$ using the CLIP model with ViT-B/32, similar to VPT~\cite{pmp:vpt} and CoOp~\cite{pmp:coop}.
In prompt-based contrastive learning, we adopt various contrastive learning schemes including augmentation-driven~\cite{drl:curl, drl:atc, drl:cody} and behavior-driven~\cite{drl:aco, drl:adat, drl:pse, drl:stacore} contrastive learning, where the prompt length sets to be 8.
In policy learning, we exploit online learning (i.e., PPO~\cite{rl:ppo}) for AI2THOR and imitation learning (i.e., DAGGER~\cite{rl:dagger}) for egocentric-Metaworld and CARLA.

\begin{table*}[h]
\centering
\caption{Zero-shot Performance. The policies of each method ($\ourmodel$ and the baselines) are learned on 4 source domains. The \textit{Source} column presents the performance for those source domains. In all evaluations, we use 30 seen target domains and 10 unseen target domains. The \textit{Seen Target} column presents the performance for the seen target domains, and the \textit{Unseen Target} column presents the performance for the unseen target domains. The unseen target domains are not used for representation learning.}
\label{tab:main_pl}
\begin{subtable}[Zero-shot Performance in AI2THOR with Object and Point Goal Navigation Tasks]{
\label{tab:main_pl:ai2thor}
\adjustbox{max width=0.98\linewidth}{
    \begin{tabular}{l ccc ccc}
    \toprule
    \multirow{2}{*}{Method}
    & \multicolumn{3}{c}{ObjectNav.} & \multicolumn{3}{c}{PointNav.} \\ 
    \cmidrule(rl){2-4} \cmidrule(rl){5-7}
     & Source & Seen Target & Unseen Target & Source & Seen Target & Unseen Target \\
    \midrule
    LUSR
    & $53.3{\pm1.1}$
    & $21.3{\pm1.9}$
    & $15.1{\pm1.8}$

    & $85.6{\pm4.6}$
    & $71.8{\pm3.8}$
    & $62.4{\pm5.8}$ \\

    CURL
    & $51.3{\pm1.0}$
    & $8.0{\pm0.1}$
    & $6.9{\pm1.3}$

    & $70.8{\pm7.4}$
    & $55.2{\pm2.7}$
    & $54.8{\pm3.0}$ \\

    ATC
    & $82.2{\pm9.7}$
    & $72.3{\pm3.3}$
    & $51.3{\pm8.6}$

    & $95.0{\pm3.3}$
    & $89.1{\pm1.9}$
    & $81.9{\pm3.6}$ \\

    ACO
    & $55.0{\pm23.8}$
    & $39.6{\pm21.5}$
    & $35.8{\pm5.8}$

    & $91.1{\pm6.3}$
    & $73.4{\pm2.0}$
    & $67.5{\pm2.8}$ \\
    
    EmbCLIP
    & $89.3{\pm3.0}$
    & $77.6{\pm1.3}$
    & $59.0{\pm6.4}$

    & $95.3{\pm4.6}$
    & $84.5{\pm1.9}$
    & $77.4{\pm1.4}$ \\
    
    $\ourmodel$
    & $\mathbf{96.3{\pm1.0}}$
    & $\mathbf{83.3{\pm0.3}}$
    & $\mathbf{79.7{\pm6.4}}$

    & $\mathbf{97.8{\pm1.0}}$
    & $\mathbf{89.7{\pm1.6}}$
    & $\mathbf{84.3{\pm2.0}}$ \\
    \bottomrule
    \end{tabular}
}}
\end{subtable}
\begin{subtable}[Zero-shot Performance in egocentric-Metaworld with Reach and Reach-wall Tasks]{
\label{tab:main_pl:metaworld}
\adjustbox{max width=0.98\linewidth}{
    \begin{tabular}{l ccc ccc}
    \toprule
    \multirow{2}{*}{Method}
    & \multicolumn{3}{c}{Reach} & \multicolumn{3}{c}{Reach-Wall} \\ 
    \cmidrule(rl){2-4} \cmidrule(rl){5-7}
     & Source & Seen Target & Unseen Target & Source & Seen Target & Unseen Target \\
    \midrule
    LUSR
    & $100.0{\pm0.0}$
    & $46.0{\pm15.1}$
    & $44.7{\pm2.3}$

    & $50.0{\pm10.0}$
    & $33.3{\pm6.1}$
    & $30.7{\pm6.4}$ \\

    CURL
    & $100.0{\pm0.0}$
    & $53.3{\pm5.0}$
    & $46.7{\pm3.1}$

    & $43.3{\pm15.3}$
    & $2.0{\pm0.0}$
    & $0.7{\pm1.2}$ \\

    ATC
    & $100.0{\pm0.0}$
    & $71.3{\pm8.1}$
    & $72.0{\pm2.0}$

    & $66.7{\pm5.8}$
    & $5.3{\pm1.2}$
    & $4.0{\pm0.0}$ \\

    ACO
    & $100.0{\pm0.0}$
    & $52.0{\pm2.0}$
    & $44.0{\pm3.5}$

    & $63.3{\pm15.3}$
    & $8.7{\pm2.3}$
    & $4.7{\pm1.2}$ \\
    
    EmbCLIP
    & $100.0{\pm0.0}$
    & $64.7{\pm6.1}$
    & $66.7{\pm4.2}$

    & $\mathbf{100.0{\pm0.0}}$
    & $58.0{\pm7.2}$
    & $49.3{\pm5.0}$ \\
    
    $\ourmodel$
    & $100.0{\pm0.0}$
    & $\mathbf{88.7{\pm3.1}}$
    & $\mathbf{86.7{\pm3.1}}$

    & $\mathbf{100.0{\pm0.0}}$
    & $\mathbf{75.3{\pm3.1}}$
    & $\mathbf{67.3{\pm2.3}}$ \\
    \bottomrule
    \end{tabular}
}}
\end{subtable}
\begin{subtable}[Zero-shot Performance in CARLA with Different Maps]{
\label{tab:main_pl:carla}
\adjustbox{max width=0.99\linewidth}{
    \begin{tabular}{l ccc ccc}
    \toprule
    \multirow{2}{*}{Method}
    & \multicolumn{3}{c}{Map 1} & \multicolumn{3}{c}{Map 2} \\ 
    \cmidrule(rl){2-4} \cmidrule(rl){5-7}
     & Source & Seen Target & Unseen Target & Source & Seen Target & Unseen Target \\
    \midrule
    LUSR
    & $2141.9$
    & $635.1{\pm606.2}$
    & $1073.9{\pm212.6}$

    & $\mathbf{2279.6}$
    & $1173.7{\pm914.3}$
    & $2159.4{\pm 146.5}$ \\

    CURL
    & $ 945.4$
    & $864.2{\pm638.0}$
    & $1256.0{\pm61.6}$

    & $1050.1$
    & $1089.9{\pm824.0}$
    & $2190.3{\pm10.2}$ \\

    ATC
    & $2280.5$
    & $1684.4{\pm368.2}$
    & $1073.7{\pm618.8}$

    & $2272.2$
    & $2253.9{\pm218.7}$
    & $2200.1{\pm307.8}$ \\

    ACO
    & $2265.8$
    & $1545.6{\pm596.1}$
    & $1330.0{\pm144.5}$

    & $2270.6$
    & $2360.9{\pm88.0}$
    & $2415.5{\pm53.0}$ \\
    
    EmbCLIP
    & $2235.7$
    & $1732.2{\pm588.6}$
    & $1415.1{\pm669.9}$

    & $2262.7$
    & $2139.1{\pm 655.9}$
    & $2401.3{\pm12.3}$ \\
    
    $\ourmodel$
    & $\mathbf{2237.5}$
    & $\mathbf{1738.0{\pm163.5}}$
    & $\mathbf{1933.4{\pm29.7}}$

    & $2277.2$
    & $\mathbf{2422.5{\pm79.6}}$
    & $\mathbf{2512.9{\pm15.7}}$ \\

    \bottomrule
    \end{tabular}
}}
\end{subtable}
\end{table*}

\subsection{Zero-shot Performance}
Table~\ref{tab:main_pl} shows zero-shot performance of $\ourmodel$ and the baselines across source, seen and unseen target domains. We evaluate with 3 different seeds and report the average performance (i.e., task success rate in AI2THOR and egocentric-Metaworld, the sum of rewards in CARLA).
As shown in Table~\ref{tab:main_pl:ai2thor}, $\ourmodel$ outperforms the baselines in the AI2THOR tasks. It particularly surpasses the most competitive baseline, EmbCLIP, by achieving $5.2 \! \sim \!  5.7\%$  higher success rate for seen target domains, and $6.9 \! \sim \! 20.7\%$ for unseen target domains.
For egocentric-Metaworld, as shown in Table~\ref{tab:main_pl:metaworld}, $\ourmodel$ demonstrates superior performance with a significant success rate for both seen and unseen target domains, which is $17.3 \! \sim \! 24.0\%$ and $18.0 \! \sim \! 20.0\%$ higher than EmbCLIP, respectively.
For autonomous driving in CARLA, we take into account external environment factors, such as weather conditions and times of day, as domain factors that can influence the driving task.
In Table~\ref{tab:main_pl:carla}, $\ourmodel$ consistently maintains competitive zero-shot performance across all conditions, outperforming the baselines.

In these experiments, LUSR shows relatively low success rates, as the reconstruction-based representation model can abate some task-specific information from observations, which is critical to conduct vision-based complex RL tasks.
EmbCLIP shows the most comparative performance among the baselines, but its zero-shot performance for target domains is not comparable to $\ourmodel$.
In contrast, $\ourmodel$ effectively estimates the domain shifts pertaining to each domain factor through the use of guided attention weights, leading to robust performance in both seen and unseen target domains.
\begin{figure*}[h]
    \centering
    \subfigure[Success Rate in Seen Target Domain]{
        \centering
        \includegraphics[width=0.4\linewidth]{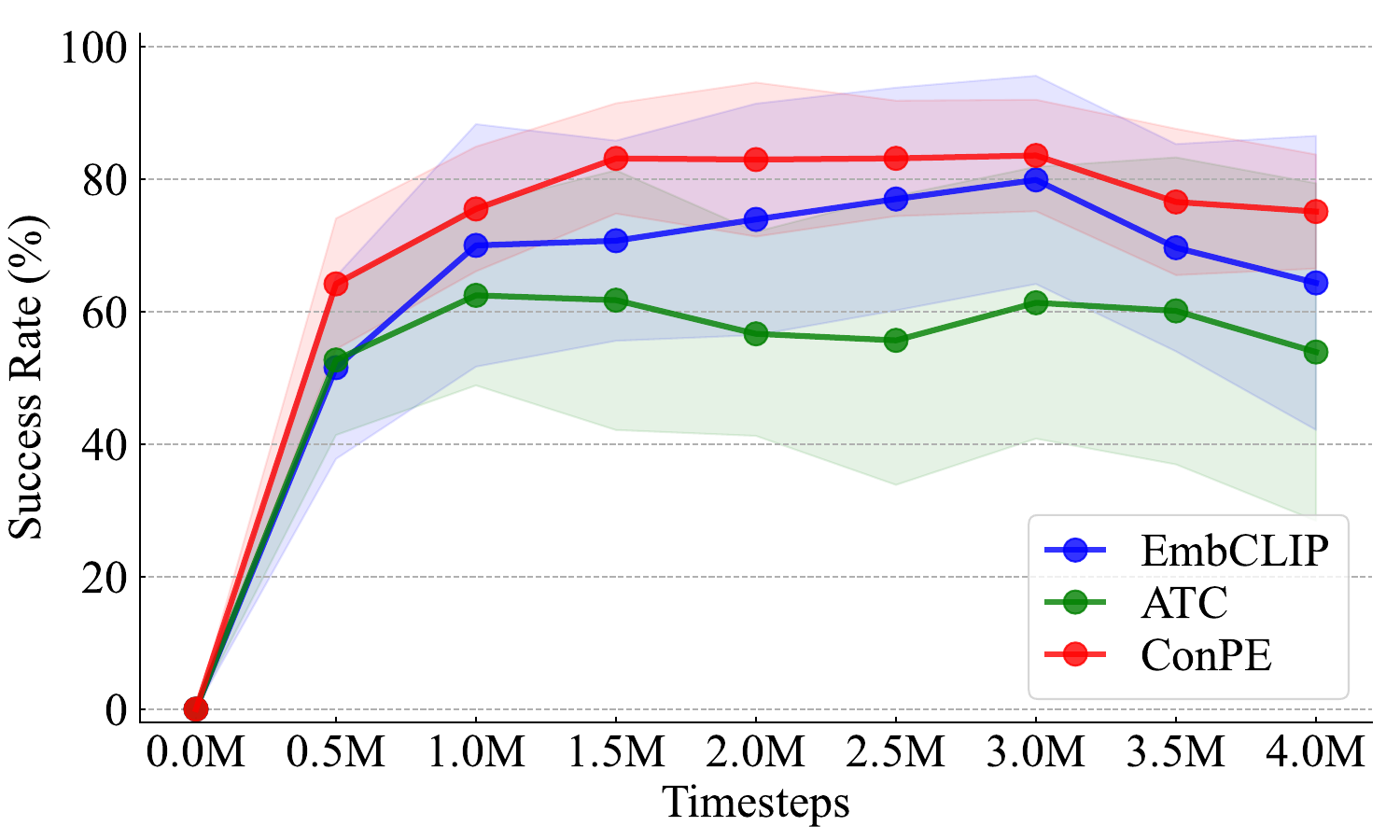}
        \label{fig:ana:efficiency_pl:train}
    }
    \hspace{10pt}
    \subfigure[Success Rate in Unseen Target Domain]{
        \centering
        \includegraphics[width=0.4\linewidth]{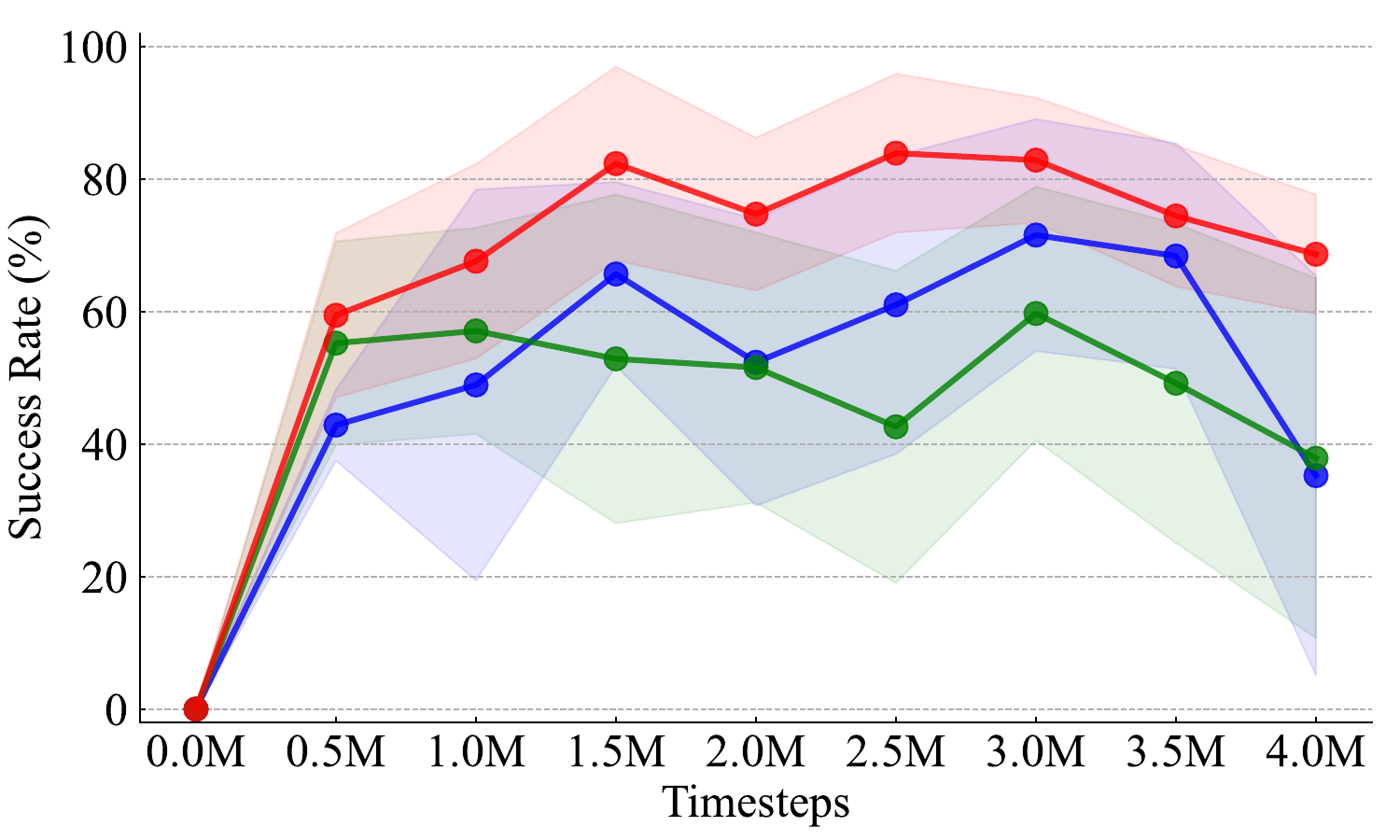}
        \label{fig:ana:efficiency_pl:test}
    }
    \caption{Sample-efficiency of Prompt Ensemble-based Policy Learning for Object Navigation in AI2THOR. The x-axis represents the number of samples (timesteps) used for policy learning, while the y-axis represents the task success rate for zero-shot evaluation.}
    \label{fig:ana}
\end{figure*}

\noindent\textbf{Sample Efficiency.}
Figure~\ref{fig:ana} presents performance with respect to samples (timesteps) that are used by $\ourmodel$ and baselines for policy learning.
Compared to the most competitive baseline EmbCLIP, $\ourmodel$ requires less than $60.0\%$ timesteps (online samples) for seen target domains and $50.0\%$ for unseen target domains to have comparable success rates.

\begin{figure*}[h]
    \centering
    \subfigure[Prompted Embeddings]{
        \centering
        \includegraphics[width=0.55\linewidth]{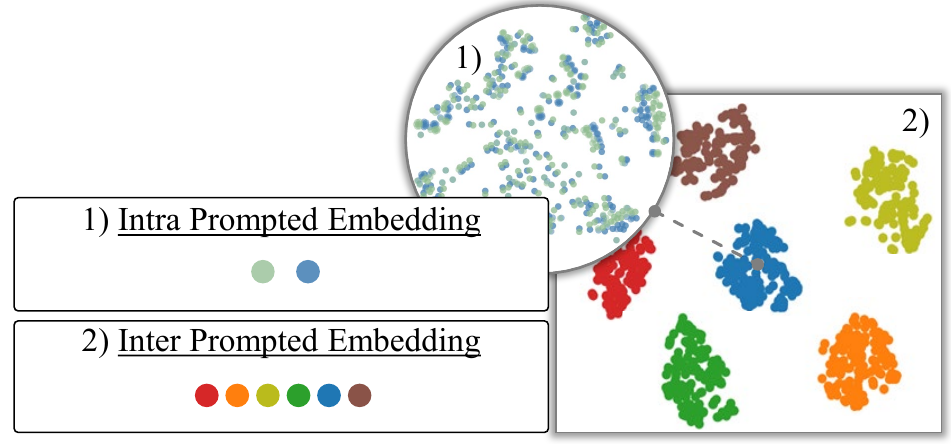}
        \label{fig:ana:interpret_emb}
    }
    \hspace{10pt}
    \subfigure[Attention Weights]{
        \centering
        \includegraphics[width=0.38 \linewidth]{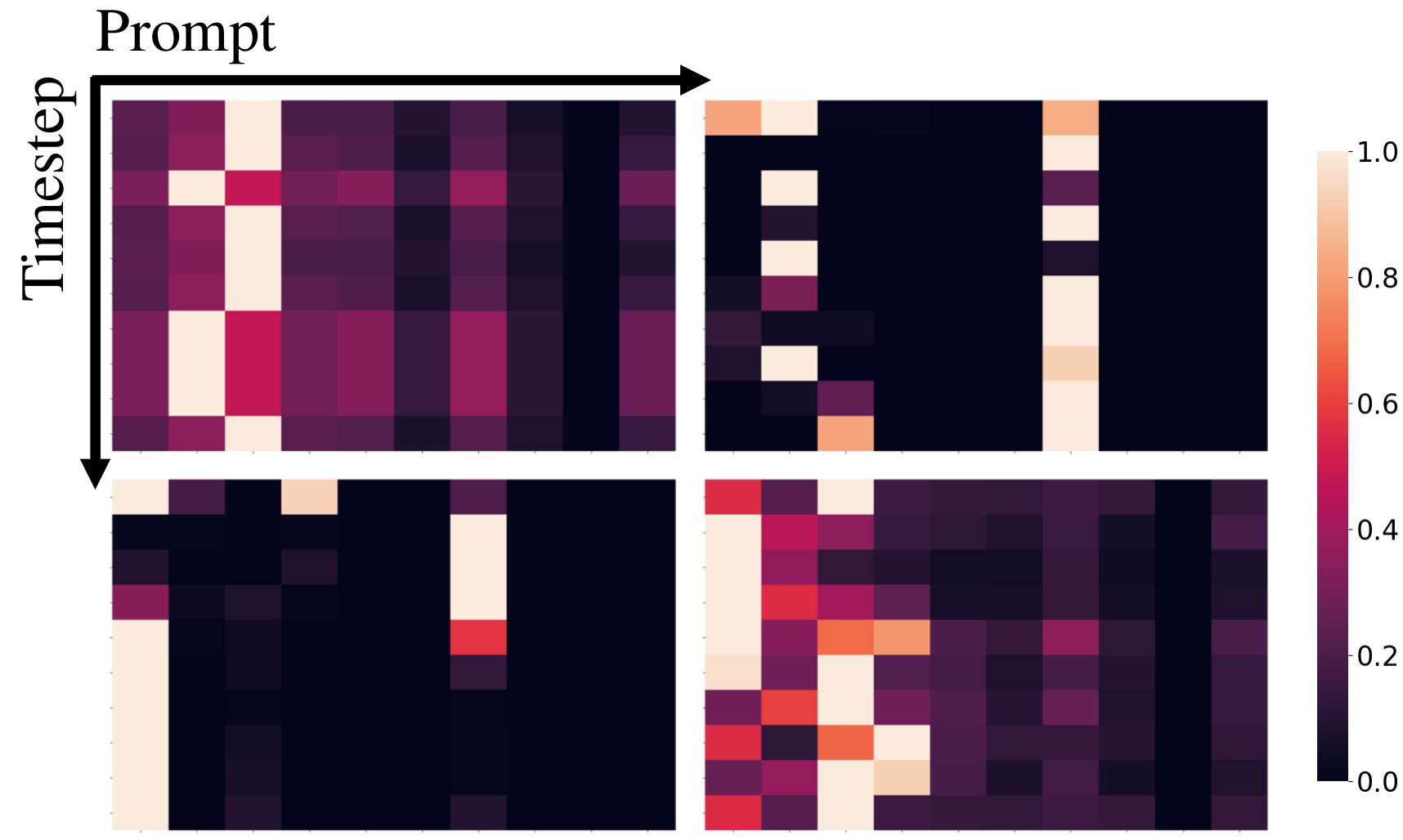}
        \label{fig:ana:interpret_attn}
    }
    \caption{Prompt Ensemble Interpretability. 
    In (a), the embeddings in the big circle are intra prompted embeddings obtained by varying domains within a domain factor, and the embeddings in the rectangle are inter prompted embeddings obtained by changing the visual prompts with aligned observation. 
    The closely located intra prompted embeddings indicate the domain-invariant knowledge, while the inter prompted embeddings clustered by different visual prompts indicate the alignment between the visual prompts and the domain factors. 
    In (b), each cell represents attention weight $\omega_i$ applied for prompted embedding $z_i$. 
    }
    \label{fig:ana:interpret}
\end{figure*}

\noindent\textbf{Prompt Ensemble Interpretability.}
Figure~\ref{fig:ana:interpret_emb} visualizes the prompted embeddings using the prompt pool obtained through $\ourmodel$.
For \textit{intra} prompted embeddings, we use observation pairs, where each pair is generated by varying domains within a domain factor.
We observe that the embeddings are paired to form the domain-invariant knowledge because the visual prompt is learned through prompt-based contrastive learning.
The \textit{inter} prompted embeddings specify that each prompt distinctly clusters prompted embeddings that correspond to the domains associated to its domain factor. 
Figure~\ref{fig:ana:interpret_attn} shows examples of the attention weight matrix of $\ourmodel$ for four different domains.
The x-axis denotes the visual prompts and the y-axis denotes timesteps.
This shows the consistency of the attention weights on the prompts across the timesteps in the same domain.

\subsection{Prompt Ensemble with a Pretrained Policy} \label{subsec:pp}
While we previously presented joint learning of a policy and the attention module $\mathcal{G}$, here we also present how to update $\mathcal{G}$ for a pretrained policy $\pi$ to make the policy adaptable to domain changes. 
In this case, we add a \textit{policy prompt} $p^v_\text{pol}$ to concentrate on task-relevant features from observations for the pretrained policy $\pi$ so that prompted embedding $\tilde{z}_0$ with the task-relevant features is incorporated into the guided-attention-based ensemble, i.e., $\pi(\mathcal{G}(\tilde{z}_0,\mathbf{z})), \ \text{where} \ \tilde{z}_0 = \mathcal{T}_\phi(o,p^v_\text{pol})$.

\begin{table*}[h]
\caption{Prompt Ensemble with a Pretrained Policy. 
The pretrained policies of each task are learned on 4 source domains. The \textit{Source} column presents the performance for those source domains. 
In all evaluations, we use 40 unseen target domains. The \textit{Target} column presents the performance for the unseen target domains not used for policy training.}
\label{tab:main_tl}
\begin{subtable}[Zero-shot Performance in AI2THOR with Visual Navigation and Room Rearrangement Tasks]{
\label{tab:main_tl:ai2thor}
\adjustbox{max width=0.99\textwidth}{
    \begin{tabular}{l cc cc cc cc}
    \toprule
    \multirow{2}{*}{Method}
    & \multicolumn{2}{c}{ObjectNav. (Aln.)} & \multicolumn{2}{c}{PointNav. (Not Aln.)} & \multicolumn{2}{c}{ImageNav. (Not Aln.)} & \multicolumn{2}{c}{RoomR. (Not Aln.)} \\ 
    \cmidrule(rl){2-3} \cmidrule(rl){4-5} \cmidrule(rl){6-7} \cmidrule(rl){8-9}
     & Source  & Target  & Source  & Target  & Souce  & Target  & Scoure  & Target  \\
    \midrule
    Pretrained
    & $87.5{\pm17.2}$
    & $65.8{\pm19.1}$
    
    & $95.3{\pm4.6}$
    & $80.9{\pm1.6}$

    & $77.2{\pm3.3}$
    & $56.2{\pm2.2}$

    & $87.3{\pm3.1}$
    & $75.2{\pm13.2}$ \\
    


    


    
    $\ourmodel$
    & $88.4{\pm1.7}$
    & $\mathbf{72.8{\pm3.1}}$
    
    & $98.9{\pm1.0}$
    & $\mathbf{84.4{\pm1.0}}$

    & $79.2{\pm1.4}$
    & $\mathbf{61.6{\pm1.1}}$

    & $93.3{\pm1.2}$
    & $\mathbf{82.2{\pm14.4}}$ \\
    \bottomrule
    \end{tabular}
}}
\end{subtable}
\begin{subtable}[Zero-shot Performance in egocentric-Metaworld with 4 Different Robot Manipulation Tasks]{
\label{tab:main_tl:metaworld}
\adjustbox{max width=0.99\textwidth}{
    \begin{tabular}{l cc cc cc cc}
    \toprule
    \multirow{2}{*}{Method}
    & \multicolumn{2}{c}{Reach (Aln.)} & \multicolumn{2}{c}{Reach-Wall (Not Aln.)} & \multicolumn{2}{c}{Button-Press (Not Aln.)} & \multicolumn{2}{c}{Door-Open (Not Aln.)} \\ 
    \cmidrule(rl){2-3} \cmidrule(rl){4-5} \cmidrule(rl){6-7} \cmidrule(rl){8-9}
     & Source  & Target  & Source  & Target  & Source  & Target  & Source  & Target  \\
    \midrule
    Pretrained
    & $100.0{\pm0.0}$
    & $65.7{\pm6.4}$

    & $100.0{\pm0.0}$
    & $58.0{\pm5.8}$
    
    & $100.0{\pm0.0}$
    & $16.8{\pm2.3}$    

    & $100.0{\pm0.0}$
    & $35.6{\pm6.2}$ \\
    
    



    


    
    $\ourmodel$
    & $100.0{\pm0.0}$
    & $\mathbf{74.7{\pm5.0}}$
    
    & $100.0{\pm0.0}$
    & $\mathbf{75.7{\pm9.0}}$
    
    & $100.0{\pm0.0}$
    & $\mathbf{73.7{\pm8.3}}$

    & $100.0{\pm0.0}$
    & $\mathbf{93.2{\pm1.1}}$ \\
    \bottomrule
    \end{tabular}
}}
\end{subtable}
\end{table*}

Table~\ref{tab:main_tl} reports zero-shot performance for the scenarios when a pretrained policy is given. 
We evaluate two different cases: \textit{aligned} (Aln.) when prompt-based contrastive learning is conducted on data from the same task of a pretrained policy; otherwise, \textit{not aligned} (Not Aln.). 
In AI2THOR, we use data from the object goal navigation task for prompt-based contrastive learning, while each pretrained policy is learned individually through one of tasks including object goal navigation, point goal navigation, image goal navigation, and room rearrangement. 
Similarly, in egocentric-Metaworld, we use data from the reach task for prompt-based contrastive learning, while each pretrained policy is learned individually through one of tasks including reach, reach-wall, button-press, and door-open.
In Table~\ref{tab:main_tl:ai2thor}, $\ourmodel$ enhances zero-shot performance of the pretrained policies by $3.5 \! \sim \! 7.0\%$ for unseen target domains in AI2THOR. This prompt ensemble adaptation requires only 400K samples, equivalent to $10\%$ of the total samples used for policy learning.
In Table~\ref{tab:main_tl:metaworld}, $\ourmodel$ significantly boosts zero-shot performance of the pretrained policies by $9.0 \! \sim \! 57.6\%$ in egocentric-Metaworld.

\subsection{Ablation Study}
Here we conduct ablation studies with AI2THOR. All the performances are reported in success rates. 

\begin{table}[!htb]
    \begin{minipage}{.42\linewidth}
      \caption{Prompt Ensemble Scalability}
      \centering
      \adjustbox{max width=\linewidth}{
      \label{tab:abl:1}
        \begin{tabular}{lccc}
        \toprule
        $n$ & Source & Seen Target & Unseen Target\\
        \midrule
        2
        & $98.7{\pm0.4}$
        & $40.5{\pm2.2}$
        & $43.0{\pm2.9}$ \\
        5
        & $96.1{\pm0.6}$
        & $59.2{\pm9.6}$
        & $45.0{\pm10.1}$ \\
        10
        & $96.3{\pm1.0}$
        & $83.3{\pm0.3}$
        & $79.7{\pm6.4}$ \\
        16
        & $91.8{\pm2.0}$
        & $83.8{\pm1.3}$
        & $77.1{\pm6.2}$ \\
        18
        & $98.5{\pm1.8}$
        & $83.3{\pm2.3}$
        & $79.0{\pm4.5}$ \\
        \bottomrule
        \end{tabular}
    }
    \end{minipage}%
    \hspace{4pt}
    \begin{minipage}{.56\linewidth}
      \centering
        \caption{Prompt Ensemble Methods}
        \adjustbox{max width=\linewidth}{
        \label{tab:abl:2}
            \begin{tabular}{l ccc}
            \toprule
            Ensemble Method & Source & Seen Target & Unseen Target\\
            \midrule
            COM-UNI-AVG
            & $52.9{\pm12.2}$
            & $51.4{\pm7.6}$
            & $43.1{\pm12.7}$ \\
            COM-WEI-AVG
            & $63.6{\pm11.2}$
            & $42.3{\pm5.2}$
            & $50.3{\pm6.1}$ \\
            ENS-UNI-AVG
            & $88.6{\pm1.4}$
            & $79.8{\pm3.4}$
            & $65.5{\pm5.0}$ \\
            ENS-WEI-AVG
            & $94.1{\pm6.3}$
            & $75.5{\pm3.5}$
            & $61.2{\pm12.7}$ \\
            $\ourmodel$
            & $\mathbf{96.3{\pm1.0}}$
            & $\mathbf{83.3{\pm0.3}}$
            & $\mathbf{79.7{\pm6.4}}$ \\
            \bottomrule
            \end{tabular}
        }
    \end{minipage} 
\end{table}

\noindent\textbf{Prompt Ensemble Scalability.}
Table~\ref{tab:abl:1} evaluates $\ourmodel$ with respect to the number of prompts ($n$).
$\ourmodel$ effectively enhances zero-shot performance for both seen and unseen target domains through prompt ensemble that captures various domain factors.
Compared to the case of $n = 2$, for $n = 10$, there was a significant improvement in zero-shot performance for both seen and unseen target domains, with increases of $42.8\%$ and $36.7\%$, respectively.
For $n \geq 10$, we observe stable performance that specifies that $\ourmodel$ can scale for combining multiple prompts to some extent.

\noindent\textbf{Prompt Ensemble Methods.} Tabel~\ref{tab:abl:2} compares the performance of various prompt integration methods~\cite{pmp:survey,attn1,attn2} including our guided attention-based prompt ensemble. 
We denote prompt-level integration as COM, and prompted embeddings-level integration as ENS.
UNI-AVG and WEI-AVG refer to uniform average and weighted average mechanisms, respectively.
$\ourmodel$ achieves superior success rates over the most competitive ensemble method ENS-UNI-AVG, showing $3.5\%$ and $14.2\%$ performance gain for seen and unseen target domains.

\begin{table}[!htb]
    \begin{minipage}{.44\linewidth}
      \caption{Prompt Ensemble Adaptation}
      \centering
        \adjustbox{max width=\linewidth}{
        \label{tab:abl:4}
        \begin{tabular}{l cc}
        \toprule
        Optimization & Source  & Target \\
        \midrule
        Pretrained
        & $95.3{\pm4.6}$
        & $80.9{\pm1.6}$ \\
        w/o $p^v_{pol}$
        & $59.5{\pm2.9}$
        & $54.2{\pm0.6}$ \\
        w $p^v_{pol}$
        & $\mathbf{98.9{\pm1.0}}$
        & $\mathbf{84.4{\pm1.0}}$\\
        E2E
        & $96.4{\pm0.5}$
        & $83.3{\pm3.3}$ \\
        \bottomrule
        \end{tabular}
        }
    \end{minipage}%
    \hspace{4pt}
    \begin{minipage}{.53\linewidth}
      \centering
        \caption{Semantic Regularized Data Augmentation}
        \adjustbox{max width=\linewidth}{
        \label{tab:abl:5}
        \begin{tabular}{l cc cc}
        \toprule
        \multirow{2}{*}{$\delta$}
          & \multicolumn{2}{c}{w/o Semantic} & \multicolumn{2}{c}{w Semantic}\\
          \cmidrule(rl){2-3} \cmidrule(rl){4-5}
          & Source & Target & Source & Target \\
        \midrule
        
        0.1
        & $97.4{\pm3.8}$
        & $83.6{\pm8.9}$
        
        & $\mathbf{100.0{\pm0.0}}$
        & $\mathbf{84.1{\pm10.2}}$ \\
        
        0.2
        & $94.7{\pm0.0}$
        & $77.6{\pm12.8}$ 
        
        & $\mathbf{94.8{\pm7.4}}$
        & $\mathbf{79.7{\pm9.1}}$ \\
        
        0.3
        & $84.2{\pm3.7}$
        & $75.3{\pm8.8}$
        
        & $\mathbf{96.1{\pm1.9}}$
        & $\mathbf{83.1{\pm10.0}}$ \\
        
        0.4
        & $80.3{\pm16.2}$
        & $74.0{\pm14.9}$
        
        & $\mathbf{86.9{\pm3.8}}$
        & $\mathbf{81.3{\pm12.3}}$ \\
        
        \bottomrule
        \end{tabular}
        }
    \end{minipage} 
\end{table}

\noindent\textbf{Prompt Ensemble Adaptation Method.} Table~\ref{tab:abl:4} shows the effect of our ensemble adaptation method for the situation when a pretrained policy is given. As explained in Section~\ref{subsec:pp}, in this situation, $\ourmodel$ can update the attention module with an additional prompt $p^v_\text{pol}$. 
Note that $p^v_{\text{pol}}$ corresponds to this case, while w/o $p^v_{\text{pol}}$ corresponds to the other case of using the attention module without  $p^v_{\text{pol}}$.
In addition, E2E denotes the fine-tuning of both the policy and the attention module along with $p^v_{\text{pol}}$. 
The results demonstrate that our method enhances the zero-shot performance of the pretrained policy, showing that $p^v_\text{pol}$ facilitates the extraction of task-specific features.

\noindent\textbf{Semantic Regularized Data Augmentation.}
So far, we have only utilized vision data, but here, we discuss one extension of $\ourmodel$ using semantic information.
Specifically, we use a few samples of object-level text descriptions to regularize the data augmentation process in policy learning. This aims to mitigate overfitting issues~\cite{aug:drq, aug:s4rl}.
The detailed explanations can be found in Appendix. 
As shown in Table~\ref{tab:abl:5}, $\ourmodel$ with semantic data (w Semantic) consistently yields better performance than $\ourmodel$ without semantic data (w/o Semantic) for all noise scale settings ($\delta$). Note that the noise scale manages the variance of augmented prompted embeddings. 
This experiment indicates that $\ourmodel$ can be improved by incorporating semantic information. 

\section{Related Work}
\noindent\textbf{Adaptation in Embodied AI.}
In the literature of robotics, numerous studies focused on developing generalized visual encoders for robotic agents across various domains~\cite{majumdar2023we, nair2022r3m}, exploiting pretrained visual encoders~\cite{stone2023open, yang2023pave}, 
and establishing robust control policies with domain randomized techniques~\cite{shah2023gnm, hirose2023exaug}. 
Furthermore, in the field of learning embodied agents, a few works addressed adaptation issues of agents to unseen scenarios in complex environments, using data augmentation techniques~ \cite{emb:speaker,emb:take,emb:dialfred,emb:vision,emb:envedit} or adopting self-supervised learning schemes~\cite{emb:rein,emb:explore,emb:moda}.
Recently, several works showed the feasibility and benefits of adopting large-scale pretrained vision-language models for embodied agents~\cite{emb:embclip,emb:clipnav,emb:clipwheel,emb:lmnav}.
Our work is in the same vein of these prior works of embodied agents, but unlike them, we explore visual prompt learning and ensembles, aiming to enhance both zero-shot performance and sample-efficiency.

\noindent\textbf{Decoupled RL Structure.}
The decoupled structure, where a state representation model is separated from RL, has been investigated in vision-based RL~\cite{drl:darla,drl:curl,drl:lusr}. 
Recently, contrastive representation learning on expert trajectories gains much interest, as it allows expert behavior patterns to be incorporated into the state encoder even when a policy is not jointly learned~\cite{drl:atc,drl:adat}. They established generalized state representations, yet in that direction, sample-efficiency issues in both representation learning and policy learning remain unexplored.  

\noindent\textbf{Prompt-based Learning.} Prompt-based learning or prompt tuning is a parameter-efficient optimization method for large pretrained models. 
Prompt tuning was used for computer vision tasks, optimizing a few learnable vectors in the text encoder~\cite{pmp:coop}, and it was also adopted for vision transformer models to handle a wide range of downstream tasks~\cite{pmp:vpt}. 
Recently, visual prompting~\cite{pmp:vpclip} was introduced, and
both visual and text prompt tuning were explored together in the multi-modal embedding space~\cite{pmp:maple,pmp:upt}. 
We also use visual prompt tuning, but we concentrate on the ensemble of multiple prompts to tackle complex embodied RL tasks. We take advantage of the fact that the generalized representation capability of different prompts can vary depending on a given task and domain, and thus we strategically utilize them to enable zero-shot adaptation of RL policies.

\section{Conclusion}
\noindent\textbf{Limitation.} Our $\ourmodel$ framework exploits visual inputs and their relevant domain factors for policy adaptation. For environments where domain changes extend beyond those domain factors, the adaptability of the framework might be constrained. 
In our future work, we will adapt the framework with semantic knowledge based on pretrained language models to improve the policy generalization capability for embodied agents in dynamic complex environments and to cover various scenarios associated with multi-modal agent interfaces.

\noindent\textbf{Conclusion.}
In this work, we presented the $\ourmodel$ framework, a novel approach that allows embodied RL agents to adapt in a zero-shot manner across diverse visual domains, exploring the ensemble structure that incorporates multiple contrastive visual prompts. The ensemble facilitates domain-invariant and task-specific state representations, thus enabling the agents to generalize to visual variations influenced by specific domain factors. 
Through various experiments, we demonstrated that the framework can enhance  policy adaptation across various domains for vision-based object navigation, rearrangement, manipulation tasks as well as autonomous driving tasks.  

\section{Acknowledgement}
We would like to thank anonymous reviewers for their valuable comments and suggestions. This work was supported by Institute of Information \& communications Technology Planning \& Evaluation (IITP) grant funded by the Korea government (MSIT) (No. 2022-0-01045, 2022-0-00043, 2020-0-01821, 2019-0-00421) and by the National Research Foundation of Korea (NRF) grant funded by the MSIT (No. NRF-2020M3C1C2A01080819, RS-2023-00213118).

\printbibliography

\newpage
\appendix
\title{Appendix}

\section{Prompt-based Contrastive Learning}
To construct domain-invariant representations efficiently for egocentric perception data in embodied agent environments, we adopt contrastive tasks for visual prompt learning on the domain factors, which can be learned from a few expert demonstrations.

\begin{wrapfigure}{R}{.54\linewidth}
    \centering
    \includegraphics[width=.52\textwidth]{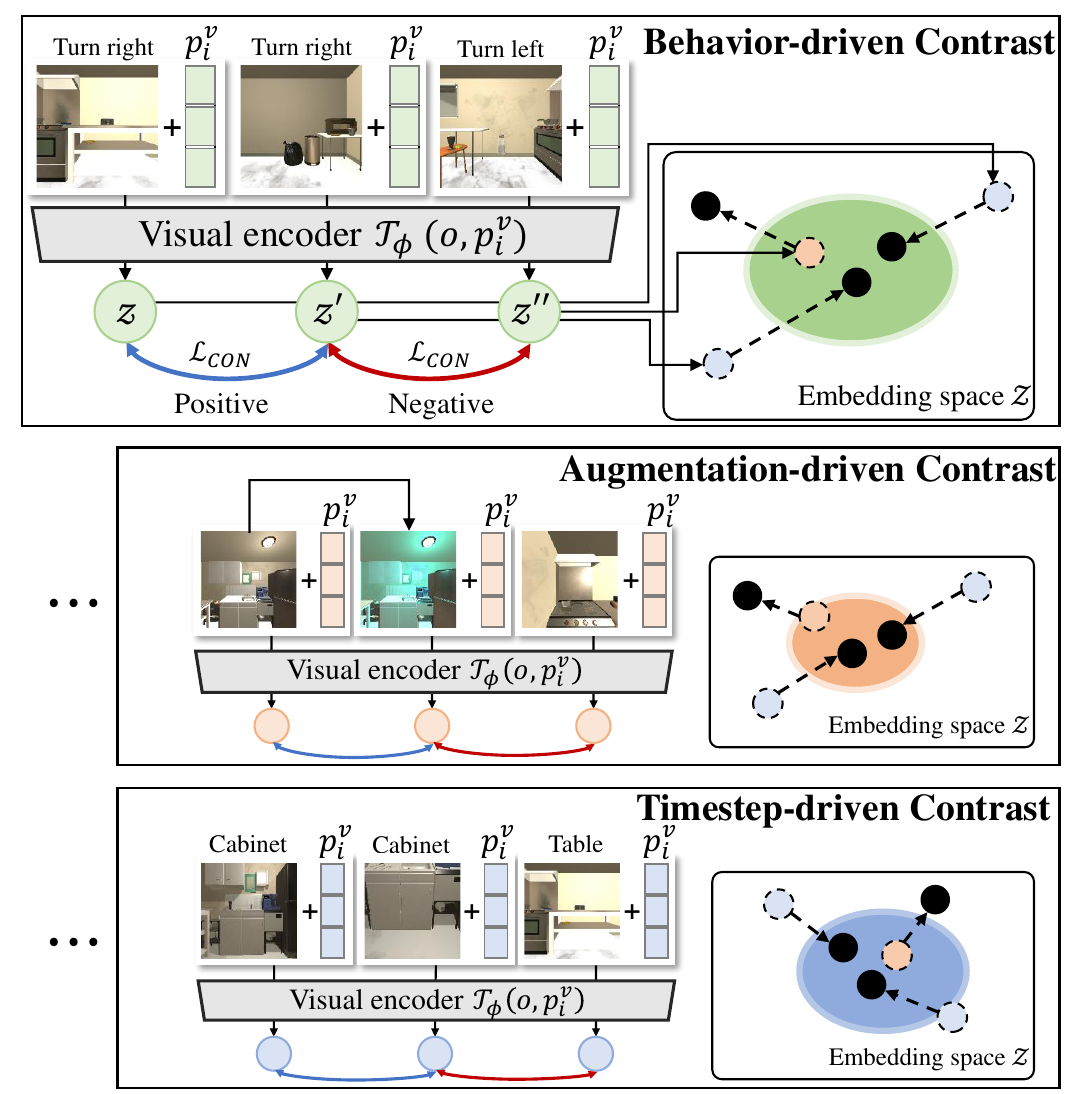}
    \caption{Prompt-based Contrastive Learning with Different Contrastive Tasks}
    \label{sub:fig:contrast}
\end{wrapfigure}

When conducting prompt-based contrastive learning, as shown in Figure~\ref{sub:fig:contrast} and explained below, we specifically adopt several methods to generate positive visual observation pairs for different domain factors. 

Consider a pretrained model $\mathcal{T}_{\phi}$ parameterized by $\phi$ that maps observations $o \in \Omega$ to the embedding space $\mathcal{Z}$. 
The contrast function $P: \Omega \times \Omega \rightarrow \{0, 1\}$ discriminates whether an observation pair is positive or not.
Then, we fine-tune $\mathcal{T}_{\phi}$ by learning a visual prompt $p^v$ through contrastive learning, 
where the contrastive loss function~\cite{con:coding} is defined as Equation (2) in the main manuscript.

\noindent\textbf{Behavior-driven Contrast.} Similar to~\cite{drl:adat}, we exploit expert actions to obtain positive sample pairs from expert trajectories of different domains. 
With observation and action pairs $(o, a), (o', a')$, a behavior-driven contrast function is defined as $P_{\text{beh}}(o, o') = \mathbbm{1}_{a = a'}$.
If the environment has a discrete action space, the behavior-driven contrast can be applied immediately to obtain positive sample pairs; otherwise, it can be applied after discretizing continuous actions with unsupervised clustering algorithms such as $k$-means clustering~\cite{kmeans}. 

\noindent\textbf{Augmentation-driven Contrast.} 
Similar to visual domain randomization techniques~\cite{con:simclr, con:hansen}, we use data augmentation (e.g., color perturbation~\cite{aug:augmentation}) for unstructured pixel-level visual domain factors such as illumination. 
An augmentation-driven contrast function is defined as $P_{\text{aug}}(o, o') = \mathbbm{1}_{o' = AUG(o)}$, where $AUG$ augments $o$.

\noindent\textbf{Timestep-driven Contrast.} Similar to~\cite{drl:atc}, we exploit timesteps of expert trajectory to obtain positive sample pairs across different domains.
With observation and timestep pairs $(o, t), (o', t')$, a timestep-driven contrast function is defined as $P_{\text{tim}}(o, o') = \mathbbm{1}_{t = t'}$, where $t - k \leqq t' \leqq t + k$ and $k$ is hyperparameter.

\begin{figure}[h]
\centering
\includegraphics[width=0.92\linewidth]{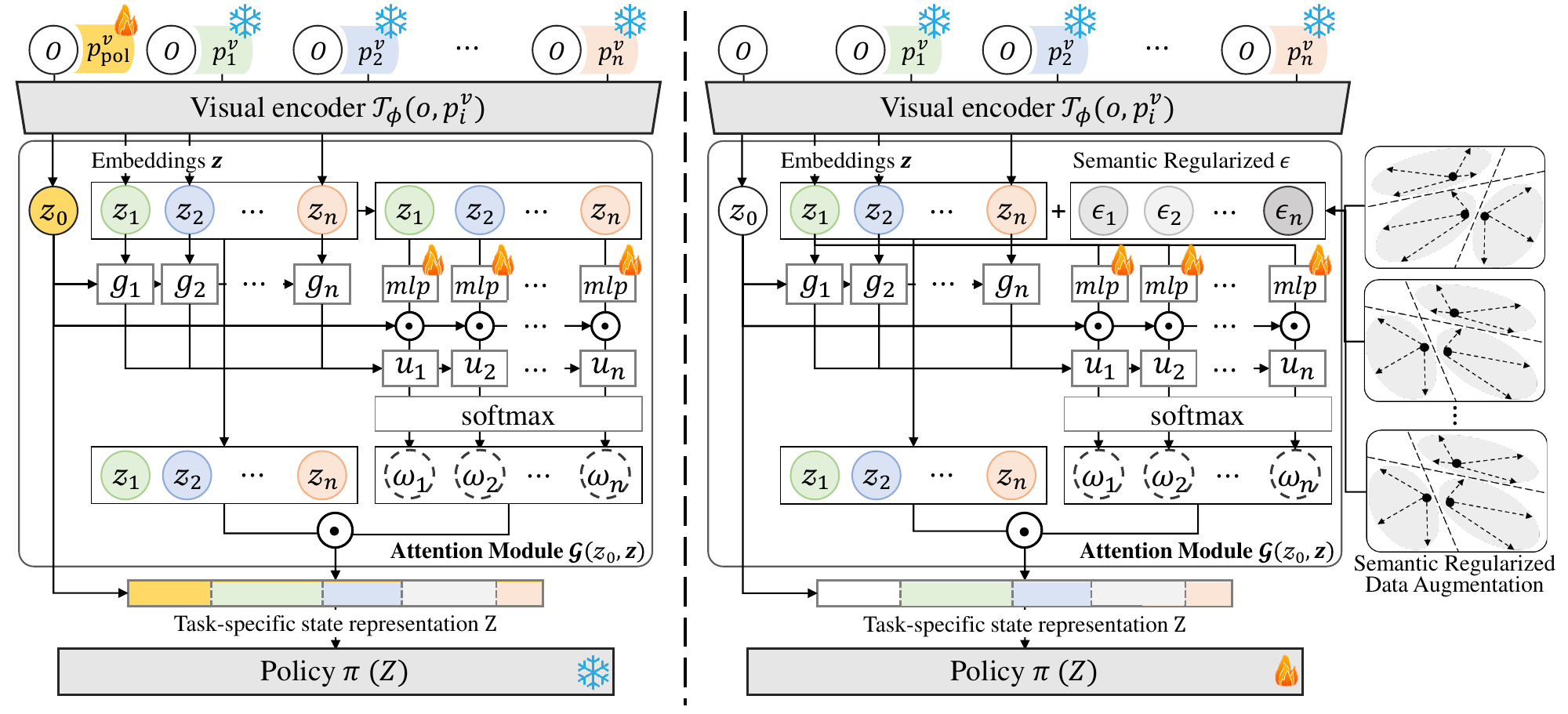}
\caption{Guided-Attention-based Prompt Ensemble. The cosine similarity-guided attention module $\mathcal{G}$ generates task-specific state representations by combining multiple prompted embeddings, and it is learned with a policy network $\pi$. The left part of the figure illustrates prompt ensemble adaptation to a pretrained policy, while the right part shows the semantic regularized data augmentation scheme.} 
\label{sub:fig:Method}
\end{figure}

\section{Prompt Ensemble with a Pretrained Policy}
As mentioned in the main manuscript, we devise an optimization method to update $\mathcal{G}$ specifically for a pretrained policy $\pi$. 
Specifically, we use a \textit{policy prompt} $p^v_\text{pol}$ that focuses on task-relevant features from observations for  $\pi$. By incorporating the prompted embedding $\tilde{z}_0$, which contains these task-relevant features, into the guided-attention-based ensemble, we can effectively integrate the policy $\pi$ with the attention module, resulting in $\pi(\mathcal{G}(\tilde{z}_0,\mathbf{z}))$. Here, $\tilde{z}_0$ is obtained by applying the transformation to the observation $o$ using the policy prompt $p^v_\text{pol}$.

As such, $\ourmodel$ enables efficient adaptation of the attention module to a pretrained policy by fine-tuning only a small number of parameters. This achieves robust zero-shot performance for unseen domains in different tasks.

Algorithm~\ref{sub:alg:framework_pretrained} shows the procedures of $\ourmodel$ to adapt the attention module for a pretrained policy. The first half corresponds to prompt-based contrastive learning and the other half corresponds to learning of the attention module $\mathcal{G}$ with a pretrained policy $\pi(Z)$. 
This is slightly extended from the algorithm in the main manuscript, where joint learning of the attention module and a policy is explained.  

\begin{algorithm}[h]
\caption{Procedure in $\ourmodel$ with a Pretrained Policy}
\label{sub:alg:framework_pretrained}
Dataset $\mathcal{D}= \{(o_1, o'_1), ...\}$, replay buffer $Z_D \leftarrow \emptyset$, pretrained vision-language model $\mathcal{T}_\phi$ \\
Visual prompt pool $\mathbf{p}^v = [p^v_1,...,p^v_n]$, attention module $\mathcal{G}$ \\
Pretrained policy $\pi$ 
\begin{algorithmic}[1] 
\STATE \textit{/* Prompt-based Contrastive Learning */}
\FOR{$i = 1, ..., n$}
    \WHILE{not converge}
        \STATE{Sample a batch $\mathcal{B}_{P_i} = \{(o_{j},o_{j}')\}_{j\leq m} \sim \mathcal{D}$}
        \STATE{Update prompt $p^v_i \gets p^v_i - \nabla\mathcal{L}_{\text{CON}}(p^v_i, \mathcal{B}_{P_i})$ using main manuscript Equation (3)}
    \ENDWHILE
\ENDFOR
\STATE \textit{/* Prompt Ensemble Learning with a Pretrained Policy */} \\
\FOR{each environment step}
    \STATE{Sample action $a = \pi(\mathcal{G}(\mathcal{T}_{\phi}(o), \textbf{z}))$ using main manuscript Equation (5), (6)}
    \STATE{Store $Z_D \gets Z_D \cup \{(\textbf{z}, a, r)\}$}
    \STATE{Optimize module $\mathcal{G}$ on $\{(\textbf{z}_j, a_j, r_j)\}_{j\leq m}\sim Z_{D}$}
\ENDFOR
\end{algorithmic}
\end{algorithm}

\section{Semantic Regularized Data Augmentation}
For a source environment that is sufficiently accessible, the attention module and policy network can be jointly trained by RL algorithms. In this case, to address overfitting problems to the source environment, data augmentation methods can be adopted. 
For example, when training a policy, it is feasible to add Gaussian noise to each prompted embedding as part of data augmentation to avoid overfitting~\cite{aug:drq, aug:s4rl}.

To enhance both policy optimization and zero-shot performance, we investigate semantic regularization schemes in the CLIP embedding space, which are specific to the prompt ensemble-based policy learning.
Specifically, using a few object-level descriptions in datasets, we  control the noise effectively.

In our semantic regularization, the language prompt $p^l_i  = [e^l_1, e^l_2,...,e^l_{u'}],\ e^l_i \in \mathbb{R}^{d'}$ is pretrained with description data and fixed $p^v_i$.  Then, $p^l_i$ is used as a semantic regularizer, where $e^l_i$ is a continuous learnable vector of the word embedding dimension $d'$ (e.g., 512 for CLIP language encoder) and $u'$ is the length of a language prompt.
Similar to~\cite{pmp:coop}, we adopt language prompt learning schemes. We also use the binary cross entropy loss for observations and object-level description pairs $(o, m)$,
\begin{equation}
    \mathcal{L}_{\text{BCE}}(p^v_i, p^l_i) = \sum_{k = 1}^n \sum_{q \in \Phi} [\log f(\mathcal{T}_\phi(o_k, p^v_i), \mathcal{T}_\phi(q, p^l_i))
    -\mathbbm{1}_{q\in m_k} \log f(\mathcal{T}_\phi(o_k, p^v_i), \mathcal{T}_\phi(q, p^l_i)) ]
\label{loss:LanguageContrastiveLoss}
\end{equation}
where $\Phi$ is the collection of object-level descriptions and $f$ is a cosine similarity function.

Given representation $z_i = \mathcal{T}_\phi(o, p^v_i)$,
we add a small Gaussian noise $\epsilon \sim \mathcal{N}(0, \delta)$ with variance $\delta$ to $z_i$. 
For all object-level descriptions $q_i$ in the given $\Phi$, we maintain regularizing augmented representations to hold the below condition.
\begin{equation}
    \text{CL}(z_i+\epsilon, \mathcal{T}_\phi(q_i ,p^l_i)) = \text{CL}(z_i, \mathcal{T}_\phi(q_i ,p^l_i)), \ 
    \text{where } \text{CL}(z, z') = \mathbbm{1} _ {\{ \sigma(S( z, z')) \geq 0.5 \} }.
    \label{semantic_regularization}
\end{equation}
This tends to achieve more generalized representations for a specific domain that is relevant to the object-level descriptions, while maintaining semantic information in the representations.

Algorithm~\ref{sub:alg:framework_semantic} shows the procedure of $\ourmodel$ with semantic regularized data augmentation. This algorithm includes three steps: the first step corresponds to prompt-based contrastive learning (same as the algorithm in the main manuscript), the second step corresponds to language prompt learning (addition for this algorithm), and the third step corresponds to the modified process of joint learning for policy $\pi(Z)$ and the attention module $\mathcal{G}$ with semantic regularized data augmentation.

\begin{algorithm}[h]
\caption{Procedure of $\ourmodel$ with Semantic Regularized Data Augmentation}
\label{sub:alg:framework_semantic}
Dataset $\mathcal{D}= \{(o_1, o'_1, m), ...\}$, replay buffer $Z_D \leftarrow \emptyset$, pretrained vision-language model $\mathcal{T}_\phi$ \\
Visual prompt pool $\mathbf{p}^v = [p^v_1,...,p^v_n]$, Language prompts $p^l_1,...,p^l_n$ \\
Attention module $\mathcal{G}$, policy $\pi$ 
\begin{algorithmic}[1] 
\STATE \textit{/* Prompt-based Contrastive Learning */}
\FOR{$i = 1, ..., n$}
    \WHILE{not converge}
        \STATE{Sample a batch $\mathcal{B}_{P_i} = \{(o_{j},o_{j}')\}_{j\leq m} \sim \mathcal{D}$}
        \STATE{Update prompt $p^v_i \gets p^v_i - \nabla\mathcal{L}_{\text{CON}}(p^v_i, \mathcal{B}_{P_i})$ using main manuscript Equation (3)}
    \ENDWHILE
\ENDFOR
\STATE \textit{/* Language Prompt Learning */}
\FOR{$i = 1, ..., n$}
    \WHILE{not converge}
        \STATE{Sample a batch $\{(o_{k}, m_{k})\}_{k\leq B} \sim \mathcal{D}$}
        \STATE{Update prompt $p^l_i \gets p^l_i - \nabla\mathcal{L}_{\text{BCE}}(p^v_i, p^l_i)$ using~\eqref{loss:LanguageContrastiveLoss}}
    \ENDWHILE
\ENDFOR
\STATE \textit{/* Prompt Ensemble-based Policy Learning with Semantic Regularized Data Augmentation*/} \\
\FOR{each environment step}
    \STATE{Sample $\bm{\epsilon} = [\epsilon_1,...,\epsilon_n]$ satisfying the condition~\eqref{semantic_regularization}}
    \STATE{Compute $\mathbf{z}_{\epsilon} = \mathbf{z} + \bm{\epsilon} = [z_1 + \epsilon_1, ..., z_n + \epsilon_n]$}
    \STATE{Sample action $a = \pi(\mathcal{G}(\mathcal{T}_{\phi}(o), \textbf{z}))$ using main manuscript Equation (5), (6)}
    \STATE{Store $Z_D \gets Z_D \cup \{(\textbf{z}, a, r)\}$}
    \STATE{Jointly optimize policy $\pi$ and module $\mathcal{G}$ on $\{(\textbf{z}_j, a_j, r_j)\}_{j\leq m}\sim Z_{D}$}
\ENDFOR
\end{algorithmic}
\end{algorithm}

\section{Environments and Datasets}
\subsection{AI2THOR}
\textbf{Environment settings.} We use AI2THOR~\cite{env:ai2thor}, a large-scale interactive simulation platform for Embodied AI. 
In AI2THOR, we use iTHOR datasets that have 120 room-sized scenes with  bedrooms, bathrooms, kitchens, and living rooms. iTHOR includes over 2000 unique objects based on Unity 3D. Among embodied AI tasks in AI2THOR, we evaluate our framework with object goal navigation and point goal navigation tasks. 
We also test the image goal navigation task, a modified version of the object goal navigation, as well as the room rearrangement task for adaptation to a pretrained policy. 

\begin{table}[h]
\caption{AI2THOR Actions}
\begin{center}
\begin{small}
\begin{tabular}{ll}

\toprule
\multicolumn{2}{c}{\textbf{Actions}}\\
\midrule
\multirow{3}{*}{Navigation}         & Move {[}Ahead/Left/Right/Back{]}   \\
                                    & Rotate {[}Right/Left{]}            \\
                                    & Look {[}Up/Down{]}                   \\ 
                                    & Done        \\
\midrule
\multirow{3}{*}{\begin{tabular}[c]{@{}l@{}}Object \\ Interaction\end{tabular}} 
                                    & PickUp  {[}Object Type{]}           \\
                                    & Open {[}Object Type{]}           \\
                                    & PlaceObject                             \\ 
\bottomrule

\end{tabular}
\end{small}
\end{center}
\end{table}

\textbf{Object Goal Navigation.}
The object navigation task requires an agent to navigate through its environment and find an object of a given category (e.g., apple). 
The agent is initially placed at a random location in a near-photorealistic home, and it receives an egocentric viewpoint image and one target object instruction for each timestep. The agent uses several navigation actions such as MoveAhead and RotateRight to complete the task. 

\textbf{Point Goal Navigation.} 
In the point goal navigation task, an agent is given a specific coordinate in the environment as its target goal. Similar to the object goal navigation task, the agent receives an egocentric viewpoint image and a target coordinate for each timestep. 

\textbf{Image Goal Navigation.} In the image goal navigation task, an agent is provided with a target image that represents a desired scene configuration. The agent's goal is to navigate through the environment and reach a location where the observed scene matches the target image. 
This task involves using visual perception to compare the current scene to the target image and selects appropriate navigation actions to achieve the desired scene configuration. The agent receives egocentric viewpoint images and target image for each timestep. The task is considered successful when the agent reaches a specific position where the observed scene closely resembles the target image.

\textbf{Room Rearrangement.}
In the room Rearrangement task, the goal of an agent is to reach the goal configuration by interacting with the environment. At each timestep, the images of both the current state and goal state are given, and the agent uses navigation actions and higher-level semantic actions (e.g., PickUp CUP) to rearrange the objects and recover the goal room configuration.

\textbf{Our experiment settings.}
We adopt the similar configuration in~\cite{emb:embclip} for our experiments, while some settings are modified to evaluate zero-shot adaptation for different domains. 

We use ``FloorPlan21'' as our default environment.
For zero-shot adaptation scenarios, 75 different domains are randomly generated with several predefined domain factors (i.e., camera field of views, stride length, rotation degrees, look degrees and illuminations). 
These factors, previously examined in embodied RL studies ~\cite{chattopadhyay2021robustnav,shah2023gnm,julian2020efficient}, can be characterized by either discrete or continuous values based on their intrinsic properties. For instance, we treat rotation degrees as a discrete factor, determined based on the feasibility of task success; conversely, brightness is treated as a continuous factor, with a range extending from $0.0$ to $1.0$. Each of these domain factors is randomly selected from a uniform distribution, leading to a combination of varied domains.
Using the domain factors, we define several seen domains that can be used for representation and policy learning as well as unseen domains for evaluating the zero-shot adaptation performance.

\textbf{Datasets.} 
Using rule-based policies, we create expert datasets for contrastive representation learning. 
The datasets comprise 28,464 samples for AI2THOR.
Table \ref{tab:AI2THOR:dataset} illustrates the examples of the expert  datasets in which the experts of each domain reflect external differences amongst domains and physical differences of agents.
\textsc{SPL} is the evaluation metric, success weighted by (normalized inverse) path length~\cite{spl}. \textsc{Length} is the average episode length of entire trajectories.

\subsection{Egocentric-Metaworld}

\textbf{Environment settings.} The Metaworld benchmark~\cite{env:metaworld} includes diverse table-top manipulation tasks that require a Sawyer robot to interact with various objects. 
With different objects, such as door and button, the robot needs to manipulate them based on the object’s affordance, leading to different reward functions. At each timestep, the Sawyer robot conducts a 4-dimensional fine-grained action that determines the 3D position movements of the end-effector and the variation of gripper openness. 
For embodied AI settings, we slightly modify Metaworld to have egocentric images as observations. We use several tasks such as 
reach-v2, reach-wall-v2, button-press-topdown-v2 and door-open-v2 for our experiments.

\textbf{Reach.} In the Reach task, the objective is to control a Sawyer robot’s end-effector to reach a target position. The agent directly controls the XYZ location of the end-effector.

\textbf{Reach-Wall.} In the Reach-Wall task, the agent controls the Sawyer robot's end-effector to reach a target position in the presence of obstacles such as walls. The agent needs to plan and navigate a path that avoids collisions to the walls while reaching the target.

\textbf{Button-Press.} In the Button-Press task, the agent is required to accurately guide the Sawyer robot's end-effector to a designated button and press it. This task involves precise control and coordination to successfully interact with the button.

\textbf{Door-Open.} In the Door-Open task, the agent's objective is to manipulate a Sawyer robot's end-effector to open a door. The agent needs to grasp and manipulate the door handle to open the door.

\textbf{Our experiment settings.} 
For zero-shot adaptation scenarios, 70 different domains are randomly generated with predefined domain factors such as camera positions, gravity, wind speeds, and illuminations. Each domain factor can be represented as either discrete or continuous values, depending on its inherent nature. Regarding sampling methods, these domain factors are individually drawn from a uniform distribution to produce combinatorial domain variations.

\textbf{Datasets.} 
For predefined seen domains, we implement a rule-based expert policy to collect expert trajectory data.
The datasets comprise 3,840 samples for Metaworld.
Table \ref{tab:Metaworld:dataset} illustrates a few examples of our expert dataset where experts of each domain reflect external differences among domains and physical differences in agents. 

\subsection{CARLA}
\textbf{Environment settings.} CARLA~\cite{env:carla} is a self-driving simulation environment where an agent navigates to the target location while avoiding collisions and lane crossings. For experiments, we use the CARLA simulator v0.9.13 and choose Town10HD as our map. For RL formulation, we incorporate the RGB image data and sensor values (acceleration, velocity, angular velocity) into states, and use control steer, throttle, and brake as actions. Each action ranges from -1 to 1. The actions are automatically calibrated when the speed of the car reaches the maximum. The agent is evaluated based on the reward function below that involves the desired velocity and goal distance, 
\begin{equation}
    \mathbf{reward} = v_t \cdot \frac{v_{target}}{\|v_{target}\|} -\frac{goal\_distance}{100}
\end{equation}
where $v_t$ denotes the agent's current velocity and $v_{target}$ denotes the target velocity.  

\begin{table}[h]
\caption{CARLA Environment Configuration}
\begin{center}
\begin{small}
\begin{tabular}{cc}
\toprule
\textbf{Configures} & \textbf{Value} \\
\midrule
Observation space $\Omega$ & $[0, 1]^{224 \times 224 \times 3} \times [-1, 1]^9$ \\
Action space $\ActionSet$ & $[-1, 1]^2$\\
Maximum speed & 20km/h \\
\bottomrule
\end{tabular}
\end{small}
\end{center}
\end{table}

\textbf{Our experiment settings.} 
To implement 50 different domains, we use camera positions, camera field of views, weather conditions, times of day, and different ranges of action magnitude as domain factors. 
Each domain factor can be represented as either discrete or continuous values, depending on its inherent nature. For example, we treat weather conditions as a discrete factor, which can be classified as either clear, rainy, cloudy, or other.
Through these factors, we define several seen domains that can be used for representation and policy learning as well as unseen domains for evaluating the zero-shot adaptation performance.
For prompt-based contrastive learning, the training dataset consists of a single trajectory for each of 50 domains. In policy learning, we utilize 4 source domains across 2 different maps.

\textbf{Datasets.} 
For predefined seen domains, we implement a rule-based expert policy to collect expert trajectory data. 
The datasets comprise 7,394 samples for CARLA.
The detailed information of the expert dataset is explained in Table~\ref{tab:CARA:dataset}.

\begin{figure*}[h]
    \centering
    \subfigure[Source Domain]{
        \centering
        \includegraphics[width=0.3\linewidth]{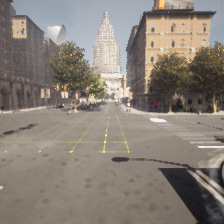}
    }
    \hspace{5pt}
    \subfigure[Seen Target Domain]{
        \centering
        \includegraphics[width=0.3\linewidth]{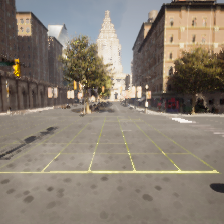}
    }
    \hspace{5pt}
    \subfigure[Unseen Target Domain]{
        \centering
        \includegraphics[width=0.3\linewidth]{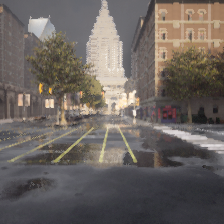}
    }
    \caption{Examples from the Source, Seen Target, and Unseen Target Domains. (a) represents the source domain, (b) depicts the seen target domain with an altered camera position, and (c) showcases the unseen target domain, differing from both the source and seen target domains.}
\end{figure*}

\section{Implementation Details}
In this section, we present the implementation details of our proposed framework $\ourmodel$ and each baseline method. $\ourmodel$ is implemented using Python v3.7, Jax v0.3.4, and PyTorch v1.13.1, and is trained on a system of an Intel(R) Core (TM) i9-10980XE processor and an NVIDIA RTX A6000 GPU.

\subsection{LUSR}
LUSR is a domain adaptation method that utilizes the latent embedding of encoder-decoder models to extract generalized representations. LUSR uses $\beta$-VAE to learn disentangled representations of different visual domains. For implementation, we use the open source (https://github.com/KarlXing/LUSR). 
When implementing LUSR, we use a CNN encoder for both DAE and $\beta$-VAE. We conduct online policy learning with PPO algorithms~\cite{rl:ppo}. The hyperparameter settings are summarized in Table~\ref{sub:hyper:LUSR}.

\begin{table*}[hbt!]
\caption{Hyperparameter Settings for LUSR}
\label{sub:hyper:LUSR}
\centering
\begin{subtable}[Hyperparameters for Representation Learning]{
\label{sub:hyper:LUSR:repr}
\adjustbox{max width=0.48\textwidth}{
\begin{tabular}{lcc cr}
\toprule
\textbf{Hyperparameters} & \textbf{Value} \\
\midrule
image size &  (3, 224, 224) RGB\\
batch size & 10 \\
train epochs & 200 \\
optimizer & Adam\\
learning rate & $1e-4$ \\
beta $\beta$ & 10 \\
latent size & 32 \\
\bottomrule
\end{tabular}
}}
\end{subtable}
\begin{subtable}[Hyperparameters for Policy Learning]{
\label{sub:hyper:LUSR:pl}
\adjustbox{max width=0.48\textwidth}{
\begin{tabular}{l cc cr}
\toprule
\textbf{Hyperparameters} & \textbf{Value} \\
\midrule
observation & (3, 224, 224) RGB \\
discount factor & 0.99\\
GAE parameter & 0.95\\
clipping parameter & 0.1\\
value loss coefficient & 0.5\\
entropy loss coefficient & 0.01\\
learning rate & $3e-4$\\
optimizer & Adam\\
training steps & 3M\\
steps per rollout & 500 \\
\bottomrule
\end{tabular}
}}
\end{subtable}
\end{table*}

\subsection{CURL and ATC}
CURL and ATC are a contrastive learning based framework for visual RL. CURL uses contrastive representation learning to extract discriminative features from raw pixels which greatly enhance sample efficiency in RL training. 
ATC enables the training of an encoder to associate pairs of observations separated by short time difference, leading to RL performance enhancement.
For implementation, we use the open source (https://github.com/facebookresearch/moco) and (https://github.com/astooke/rlpyt/tree/master/rlpyt/ul).
The hyperparameter settings of CURL and ATC are summarized in Table~\ref{sub:hyper:CURL} and ~\ref{sub:hyper:ATC}, respectively, where the other settings are the same as in Table~\ref{sub:hyper:LUSR:pl}.

\begin{table}[!htb]
    \begin{minipage}{.48\linewidth}
      \caption{Hyperparameter Settings for CURL}
      \centering
        \adjustbox{max width=\linewidth}{
        \label{sub:hyper:CURL}
        \begin{tabular}{lcccr}
        \toprule
        \textbf{Hyperparameters} & \textbf{Value} \\
        \midrule
        image size &  (3, 224, 224) RGB\\
        batch size & 256\\
        model architecture & ViT-B/32\\
        train epochs & 500\\
        optimizer & sgd\\
        learning rate & $1e-3$ \\
        \bottomrule
        \end{tabular}
        }
    \end{minipage}%
    \hspace{4pt}
    \begin{minipage}{.48\linewidth}
      \centering
        \caption{Hyperparameter Settings for ATC}
        \adjustbox{max width=\linewidth}{
        \label{sub:hyper:ATC}
        \begin{tabular}{lcccr}
        \toprule
        \textbf{Hyperparameters} & \textbf{Value} \\
        \midrule
        image size &  (3, 224, 224) RGB\\
        batch size & 256\\
        model architecture & ViT-B/32\\
        train epochs & 500\\
        optimizer & sgd\\
        timestep $k$ & 3\\ 
        learning rate & $1e-3$ \\
        \bottomrule
        \end{tabular}
        }
    \end{minipage} 
\end{table}

\subsection{ACO}
ACO utilizes augmentation-driven and behavior-driven contrastive tasks in the context of RL. For implementation, we use the open source (https://github.com/metadriverse/ACO). The hyperparameter settings are summarized in Table~\ref{sub:hyper:ACO}, where other settings are the same as in Table~\ref{sub:hyper:LUSR:pl}. 
\begin{table}[h]
\caption{Hyperparameter Settings for ACO}
\label{sub:hyper:ACO}
\begin{center}
\begin{small}
\begin{tabular}{lcccr}
\toprule
\textbf{Hyperparameters} & \textbf{Value} \\
\midrule
image size &  (3, 224, 224) RGB\\
batch size & 256\\
model architecture & ViT-B/32\\
train epochs & 500\\
optimizer & sgd\\
learning rate & $1e-3$ \\
\bottomrule
\end{tabular}
\end{small}
\end{center}
\end{table}

\subsection{EmbCLIP}
EmbCLIP is a state-of-the-art model for embodied AI tasks. By using CLIP as the visual encoder, EmbCLIP extracts generalized representation which is useful for an embodied agent, enabling the agent to effectively generalize to different environments and tasks. We use the open source (https://github.com/allenai/embodied-clip) for the implementation of AI2-THOR environments. To evaluate this with the CARLA simulator, we also use the open source (https://github.com/openai/CLIP). The configurations for policy learning are the same as in Table~\ref{sub:hyper:LUSR:pl}.

\subsection{ConPE}
The procedure of our $\ourmodel$ consists of prompt learning and policy learning steps. \\

\noindent\textbf{Prompt-based Contrastive Learning.}
In prompt learning step, for sample-efficiency, $\ourmodel$ conducts prompt-based contrastive learning, exploiting the pretrained CLIP model. We set the length of visual prompts to be 8 for each contrastive learning with Equation (2) in the main manuscript. In cases when metadata is available (the cases of using the semantic regularized data augmentation), 8 language prompts are used for~\eqref{loss:LanguageContrastiveLoss}. The hyperparameter settings are summarized in Table~\ref{sub:hyper:ConPE}. 

\begin{table*}[hbt!]
\caption{Hyperparameter Settings for $\ourmodel$'s Prompt-based Contrastive Learning}
\label{sub:hyper:ConPE}
\centering
\begin{subtable}[Augmentation-driven]{
\adjustbox{max width=0.31\textwidth}{
\begin{tabular}{lcccr}
\toprule
 \textbf{Hyperparameters} & \textbf{Value} \\
\midrule
image size &  (3, 224, 224) RGB\\
batch size & 256\\
visual prompt length & 8\\
pretrained model & CLIP ViT-B/32\\
train epoch & 1000\\
optimizer & Adam\\
learning rate & $1e-2$ \\
\bottomrule
\end{tabular}
}}
\end{subtable}
\begin{subtable}[Behavior-driven]{
\adjustbox{max width=0.31\textwidth}{
\begin{tabular}{l cc cr}
\toprule
 \textbf{Hyperparameters} & \textbf{Value} \\
\midrule
image size &  (3, 224, 224) RGB\\
batch size & 64\\
visual prompt length & 8\\
pretrained model & CLIP ViT-B/32\\
train epoch & 500\\
optimizer & Adam\\
learning rate & $1e-2$ \\
\bottomrule
\end{tabular}
}}
\end{subtable}
\begin{subtable}[Timestep-driven]{
\adjustbox{max width=0.31\textwidth}{
\begin{tabular}{l cc cr}
\toprule
 \textbf{Hyperparameters} & \textbf{Value} \\
\midrule
image size &  (3, 224, 224) RGB\\
batch size & 64\\
visual prompt length & 8\\
pretrained model & CLIP ViT-B/32\\
train epoch & 500\\
optimizer & Adam\\
timestep $k$ & 3 \\
learning rate & $1e-2$ \\
\bottomrule
\end{tabular}
}}
\end{subtable}
\end{table*}

\noindent\textbf{Prompt Ensemble-based Policy Learning.}  $\ourmodel$ obtains domain-invariant states from observations using the ensemble of multiple prompts. The prompt attention module $\mathcal{G}$ consists of as many MLP(Multi-Layer Perceptron) layers as the number of the prompts and it is learned jointly with a policy network for a given task. 
The hyperparameter settings for policy learning are the same as in Table~\ref{sub:hyper:LUSR:pl}.

\section{Additional Experiments}

\subsection{Zero-shot Performance for Seen Domains Factors}
Table~\ref{sub:tab:factors} presents the zero-shot performance of $\ourmodel$ across specific domain factors (DF). 
For example, DF0 refers to various domains where the camera position is a domain factor of interest, and LUSR's 48.5 in DF0 indicates the success rate of LUSR for the domains generated by different camera positions. 

\begin{figure}[h]
    \centering
    \includegraphics[width=0.7\linewidth]{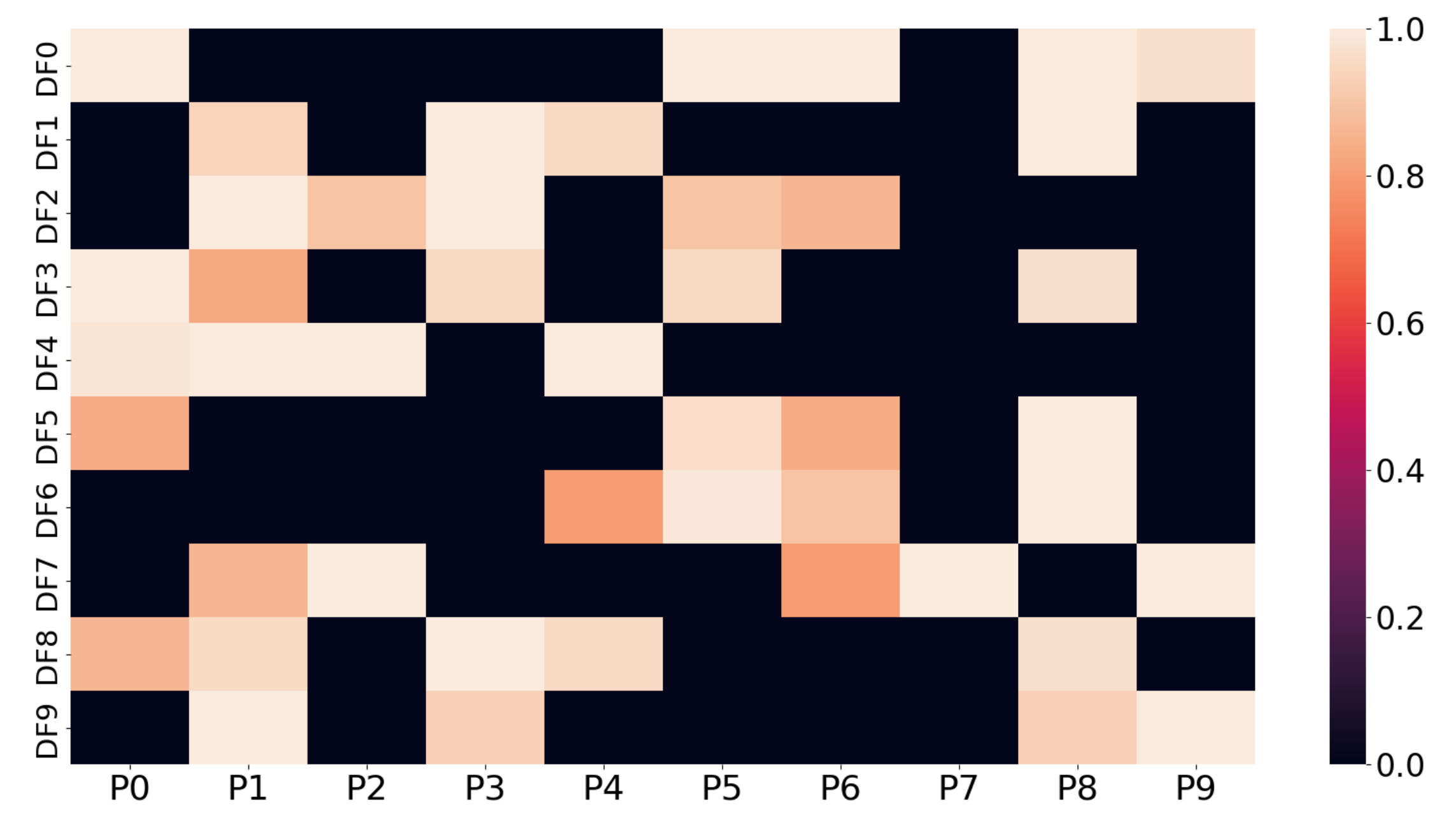}    
    \caption{Prompt Ensemble Attention Weight Matrix for Seen Domains Factors.}
    \label{sub:fig:ana:interpret}
\end{figure}

As shown, $\ourmodel$ maintains robust zero-shot performance for all the cases (DF0$\sim$ DF9), compared to the baselines. Additionally, as shown in Figure~\ref{sub:fig:ana:interpret} where a specific domain factor is changed, the attention weights assigned to the prompted embedding (denoted as P$n$, where $n=\{0..9\}$) that is trained for the corresponding domain factor are high (in the bright color).

\begin{table}[h]
\centering
\caption{
Zero-shot Performance for Seen Domains Factors
}
\label{sub:tab:factors}
\adjustbox{max width=0.99\linewidth}{
    \begin{tabular}{l cccccccccc}
    \toprule
    Method
     & DF0 & DF1 & DF2 & DF3 & DF4 & DF5 & DF6 & DF7 & DF8 & DF9\\
    \midrule
    LUSR
    & $48.5$
    & $37.2$
    & $27.5$
    & $66.4$
    & $69.5$
    & $41.6$
    & $11.7$
    & $40.6$
    & $26.8$
    & $48.1$ \\

    CURL
    & $28.6$
    & $27.3$
    & $8.4$
    & $28.3$
    & $13.0$
    & $14.9$
    & $2.2$
    & $13.2$
    & $11.8$
    & $26.6$ \\

    ATC
    & $73.2$
    & $84.9$
    & $73.1$
    & $89.8$
    & $95.4$
    & $79.8$
    & $55.8$
    & $89.8$
    & $69.0$
    & $88.1$ \\

    ACO
    & $38.3$
    & $39.6$
    & $36.0$
    & $37.2$
    & $35.9$
    & $30.8$
    & $21.5$
    & $31.1$
    & $24.1$
    & $41.1$ \\
    
    EmbCLIP
    & $71.6$
    & $79.3$
    & $83.5$
    & $92.3$
    & $96.8$
    & $\mathbf{90.8}$
    & $62.0$
    & $92.7$
    & $\mathbf{75.8}$
    & $92.4$ \\
    
    $\ourmodel$
    & $\mathbf{78.1}$
    & $\mathbf{90.9}$
    & $\mathbf{93.2}$
    & $\mathbf{95.5}$
    & $\mathbf{97.0}$
    & $88.2$
    & $\mathbf{68.7}$
    & $\mathbf{93.9}$
    & $67.0$
    & $\mathbf{93.7}$ \\
    \bottomrule
    \end{tabular}
}
\end{table}

\begin{figure*}[h]
    \centering
    \subfigure[Domain Factor 1]{
        \centering
        \includegraphics[width=0.31\linewidth]{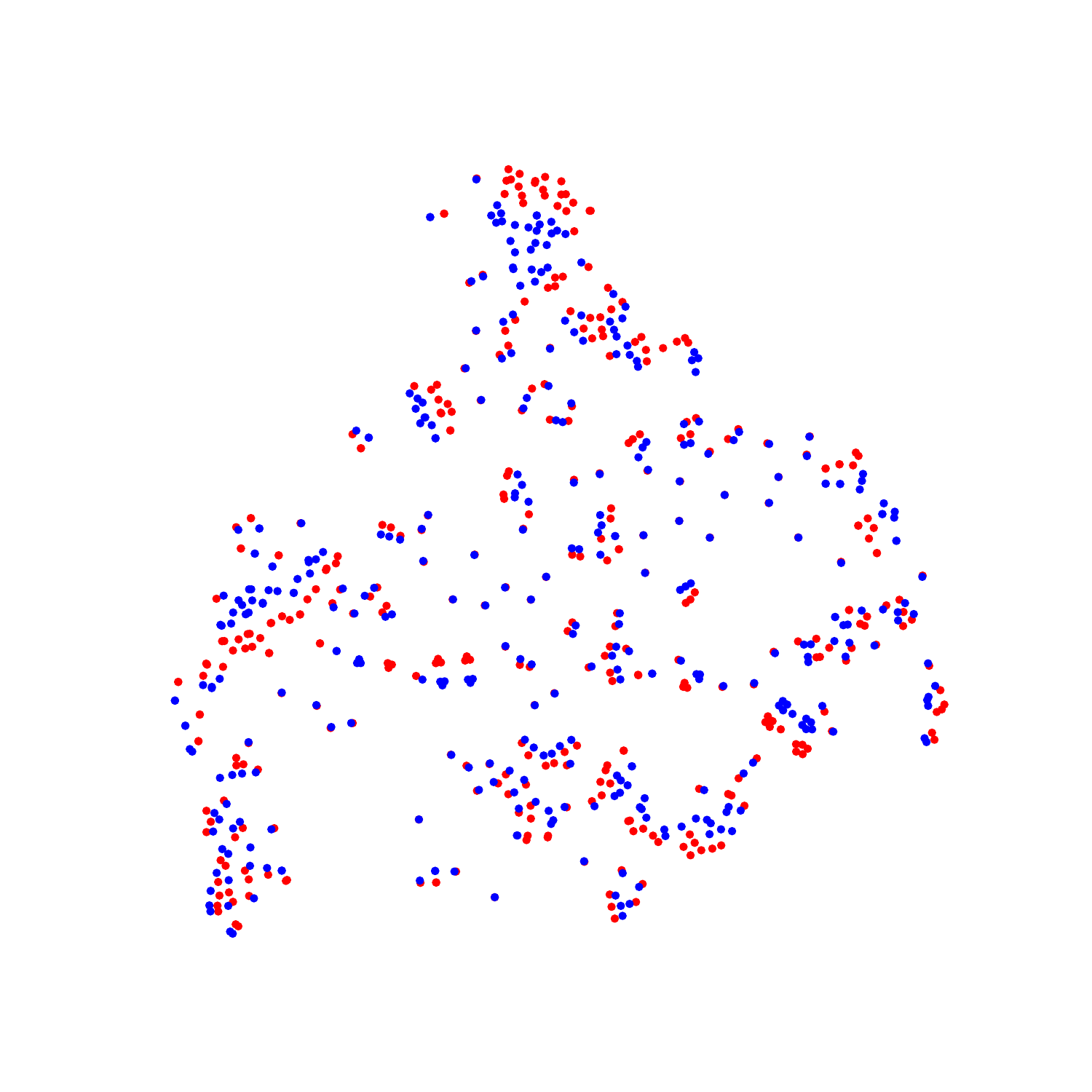}
    }
    \subfigure[Domain Factor 2]{
        \centering
        \includegraphics[width=0.31\linewidth]{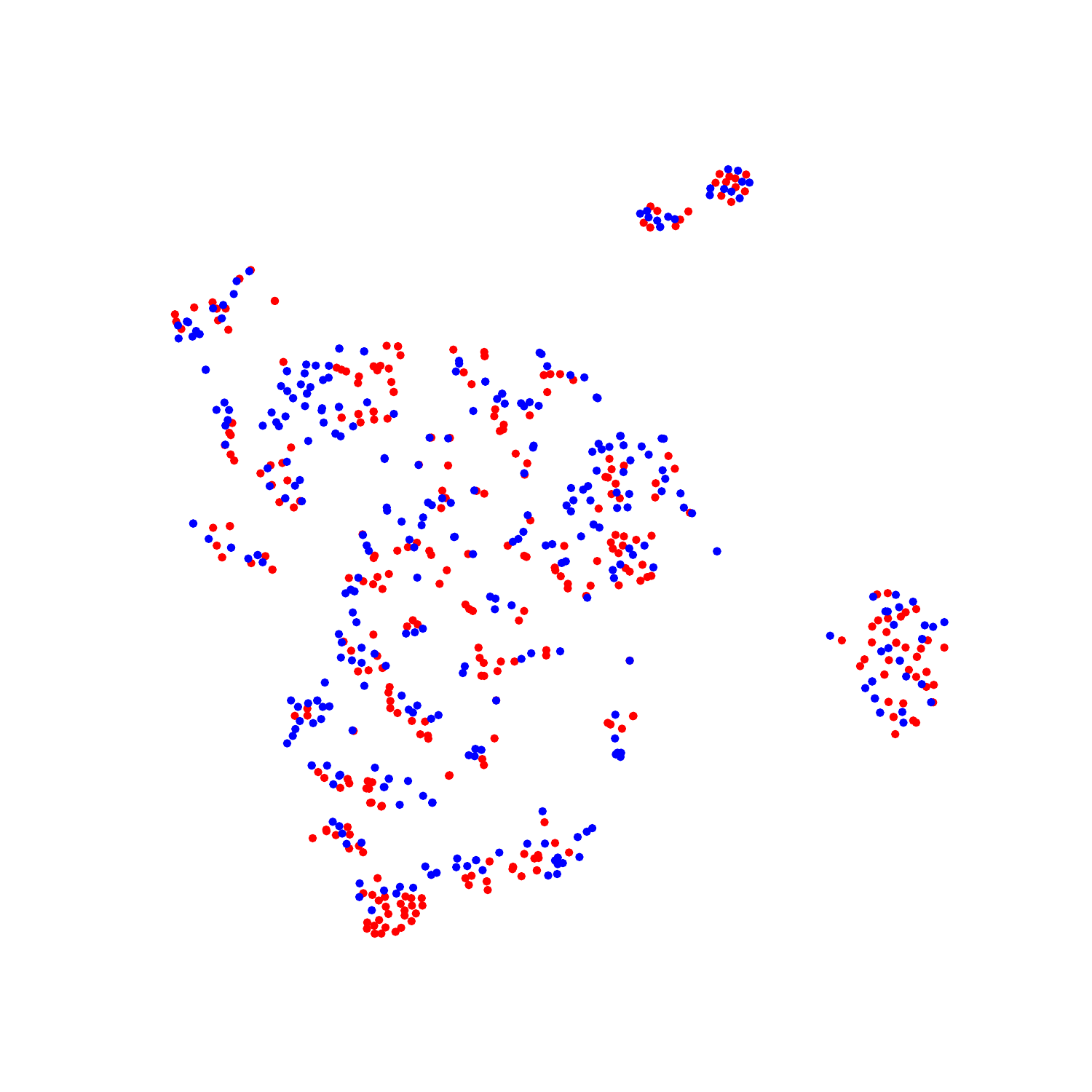}
    }
    \subfigure[Domain Factor 3]{
        \centering
        \includegraphics[width=0.31\linewidth]{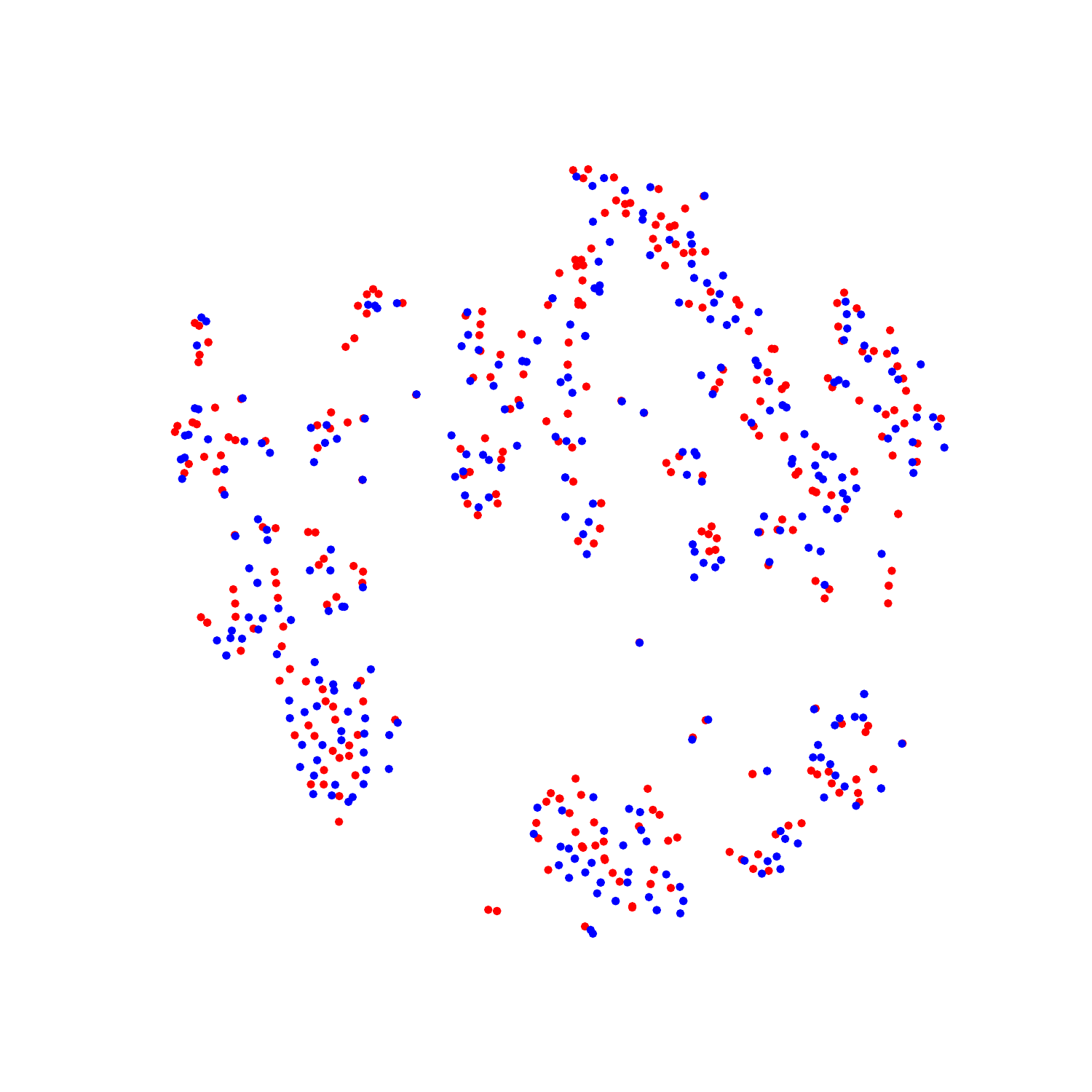}    
    }
    \caption{Intra Prompted Embeddings. Two distinct domains, represented in blue and red dots in the figures, are chosen based on different configurations of the same domain factor to assess their embedding alignment.}
\end{figure*}

\subsection{Prompt Ensemble with a Pretrained Policy}
\begin{figure*}[h]
    \centering
    \setcounter{row}{1}%
    \subfigure[Success Rate of Train Environment]{
        \centering
        \includegraphics[width=0.4\linewidth]{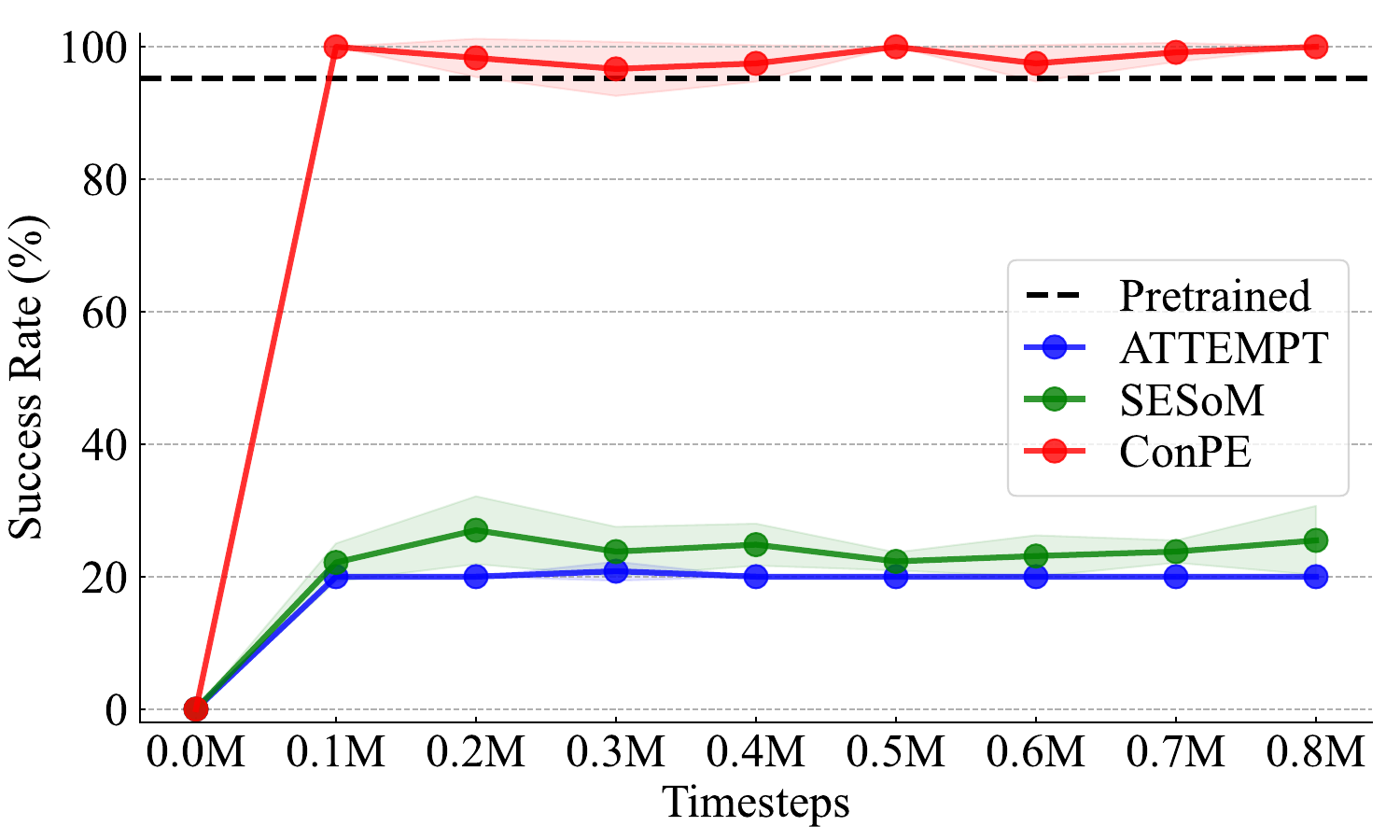}
        \label{sub:fig:ana:efficiency_tl:train}
    }
    \subfigure[Success Rate of Unseen Environment]{
        \centering
        \includegraphics[width=0.4\linewidth]{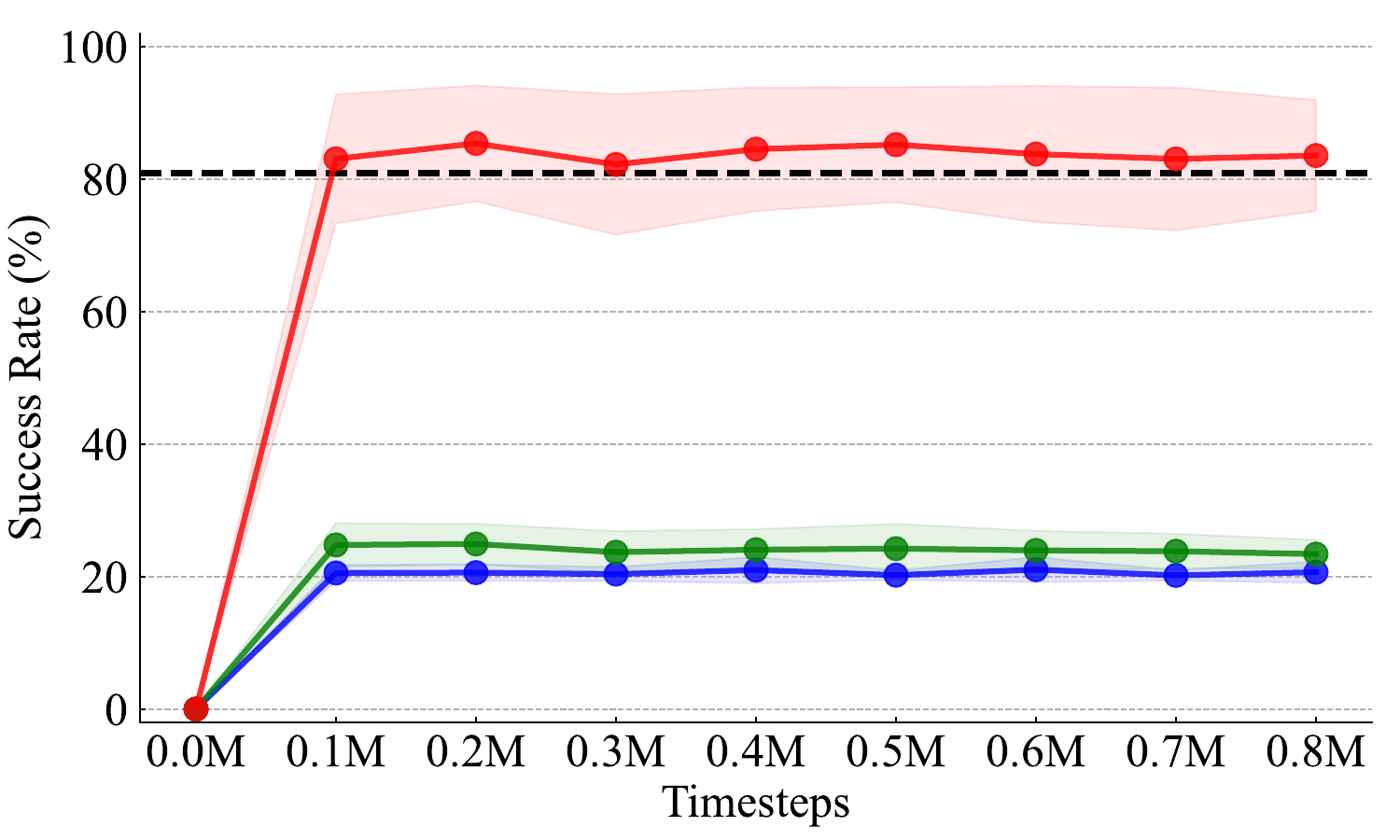}
        \label{sub:fig:tl:ana:efficiency_tl:test}
    }
    \caption{Sample-efficiency of Prompt Ensemble with a Pretrained Policy. The pretrained policy is learned on the Object Goal Navigation task and then adapted to the Point Goal Navigation task with \textit{policy prompt}.}
    \label{sub:fig:ana:efficiency_tl}
\end{figure*}

Table~\ref{tab:tl} reports detailed zero-shot performance for the scenarios when a pretrained policy is given. 
We additionally test widely used baselines that leverage large pretrained models in the fields of vision and natural language, specifically in the context of prompt-meta learning.
ATTEMPT~\cite{pmp:attempt} is a parameter-efficient multi-task language model tuning method that transfers knowledge across different tasks via a mixture of soft prompts.
SESoM~\cite{pmp:sesom} is a soft prompts ensemble method that leverages multiple source tasks and effectively improves few-shot performance of prompt tuning by transferring knowledge from the source tasks to a target task.
$\ourmodel$ demonstrates the ability to effectively improve zero-shot performance with a small number of samples, especially when a pretrained policy is given.
As shown in Figure~\ref{sub:fig:ana:efficiency_tl} (a) and (b), prompt ensemble adaptation demonstrates an increase in success rate with a small number of samples in both the train and unseen environments of the pretrained policy, compared to other baselines. This results present the sample-efficiency of our prompt ensemble adaptation method.

\begin{table*}[h]
\caption{Prompt Ensemble with a Pretrained Policy.}
\label{tab:tl}
\begin{subtable}[Zero-shot Performance in AI2THOR with Visual Navigation and Room Rearrangement Tasks]{
\label{sub:tab:tl:ai2thor}
\adjustbox{max width=0.99\textwidth}{
    \begin{tabular}{l cc cc cc cc}
    \toprule
    \multirow{2}{*}{Method}
    & \multicolumn{2}{c}{ObjectNav. (Aln.)} & \multicolumn{2}{c}{PointNav. (Not Aln.)} & \multicolumn{2}{c}{ImageNav. (Not Aln.)} & \multicolumn{2}{c}{RoomR. (Not Aln.)} \\ 
    \cmidrule(rl){2-3} \cmidrule(rl){4-5} \cmidrule(rl){6-7} \cmidrule(rl){8-9}
     & Source  & Target  & Source  & Target  & Souce  & Target  & Scoure  & Target  \\
    \midrule
    Pretrained
    & $87.5{\pm17.2}$
    & $65.8{\pm19.1}$
    
    & $95.3{\pm4.6}$
    & $80.9{\pm1.6}$

    & $77.2{\pm3.3}$
    & $56.2{\pm2.2}$

    & $87.3{\pm3.1}$
    & $75.2{\pm13.2}$ \\
    ATTEMPT
    & $2.85{\pm0.4}$
    & $3.24{\pm0.3}$
    
    & $20.2{\pm0.5}$
    & $20.7{\pm0.3}$

    & $13.8{\pm3.0}$
    & $15.0{\pm2.3}$

    & $5.3{\pm1.2}$ 
    & $3.6{\pm1.6}$ \\
    SESoM
    & $2.0{\pm6.6}$
    & $3.4{\pm5.0}$
    
    & $19.7{\pm0.5}$
    & $20.6{\pm0.1}$

    & $11.2{\pm2.4}$
    & $8.9{\pm1.2}$

    & $60.0{\pm2.0}$
    & $44.2{\pm14.0}$ \\
    
    $\ourmodel$
    & $88.4{\pm1.7}$
    & $\mathbf{72.8{\pm3.1}}$
    
    & $98.9{\pm1.0}$
    & $\mathbf{84.4{\pm1.0}}$

    & $79.2{\pm1.4}$
    & $\mathbf{61.6{\pm1.1}}$

    & $93.3{\pm1.2}$
    & $\mathbf{82.2{\pm14.4}}$ \\
    \bottomrule
    \end{tabular}
}}
\end{subtable}
\begin{subtable}[Zero-shot Performance in Egocentric-Metaworld with 4 Different Robot Manipulation Tasks]{
\label{sub:tab:tl:metaworld}
\adjustbox{max width=0.99\textwidth}{
    \begin{tabular}{l cc cc cc cc}
    \toprule
    \multirow{2}{*}{Method}
    & \multicolumn{2}{c}{Reach (Aln.)} & \multicolumn{2}{c}{Reach-Wall (Not Aln.)} & \multicolumn{2}{c}{Button-Press (Not Aln.)} & \multicolumn{2}{c}{Door-Open (Not Aln.)} \\ 
    \cmidrule(rl){2-3} \cmidrule(rl){4-5} \cmidrule(rl){6-7} \cmidrule(rl){8-9}
     & Source  & Target  & Source  & Target  & Source  & Target  & Source  & Target  \\
    \midrule
    Pretrained
    & $100.0{\pm0.0}$
    & $65.7{\pm6.4}$

    & $100.0{\pm0.0}$
    & $58.0{\pm5.8}$
    
    & $100.0{\pm0.0}$
    & $16.8{\pm2.3}$    

    & $100.0{\pm0.0}$
    & $35.6{\pm6.2}$ \\
    
    ATTEMPT
    & $73.3{\pm5.8}$
    & $33.7{\pm7.7}$
    
    & $76.7{\pm15.3}$
    & $38.0{\pm4.0}$

    & $100.0{\pm0.0}$
    & $25.7{\pm8.3}$

    & $100.0{\pm0.0}$
    & $44.0{\pm7.4}$ \\

    SESoM
    & $16.7{\pm5.8}$
    & $9.3{\pm2.1}$
    
    & $0.0{\pm0.0}$
    & $0.0{\pm0.0}$

    & $0.0{\pm0.0}$
    & $0.0{\pm0.0}$

    & $0.0{\pm0.0}$
    & $0.0{\pm0.0}$ \\
    
    $\ourmodel$
    & $100.0{\pm0.0}$
    & $\mathbf{74.7{\pm5.0}}$
    
    & $100.0{\pm0.0}$
    & $\mathbf{75.7{\pm9.0}}$
    
    & $100.0{\pm0.0}$
    & $\mathbf{73.7{\pm8.3}}$

    & $100.0{\pm0.0}$
    & $\mathbf{93.2{\pm1.1}}$ \\
    \bottomrule
    \end{tabular}
}}
\end{subtable}
\end{table*}

\begin{figure*}[h]
    \centering
    \setcounter{row}{1}%
    \subfigure[Success Rate of Train Environment]{
        \centering
        \includegraphics[width=0.4\linewidth]{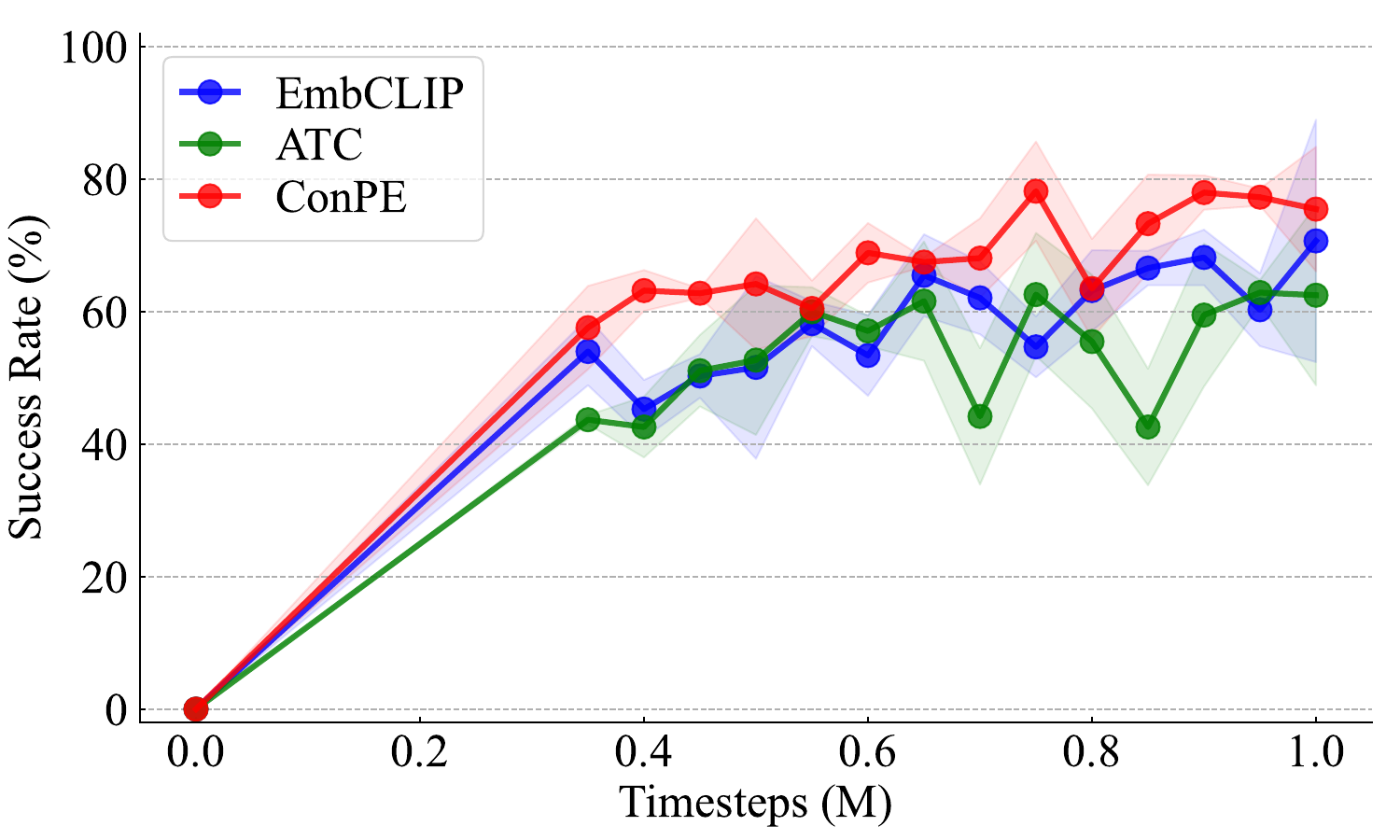}
        \label{sub:fig:efficiency:train}
    }
    \subfigure[Success Rate of Unseen Environment]{
        \centering
        \includegraphics[width=0.4\linewidth]{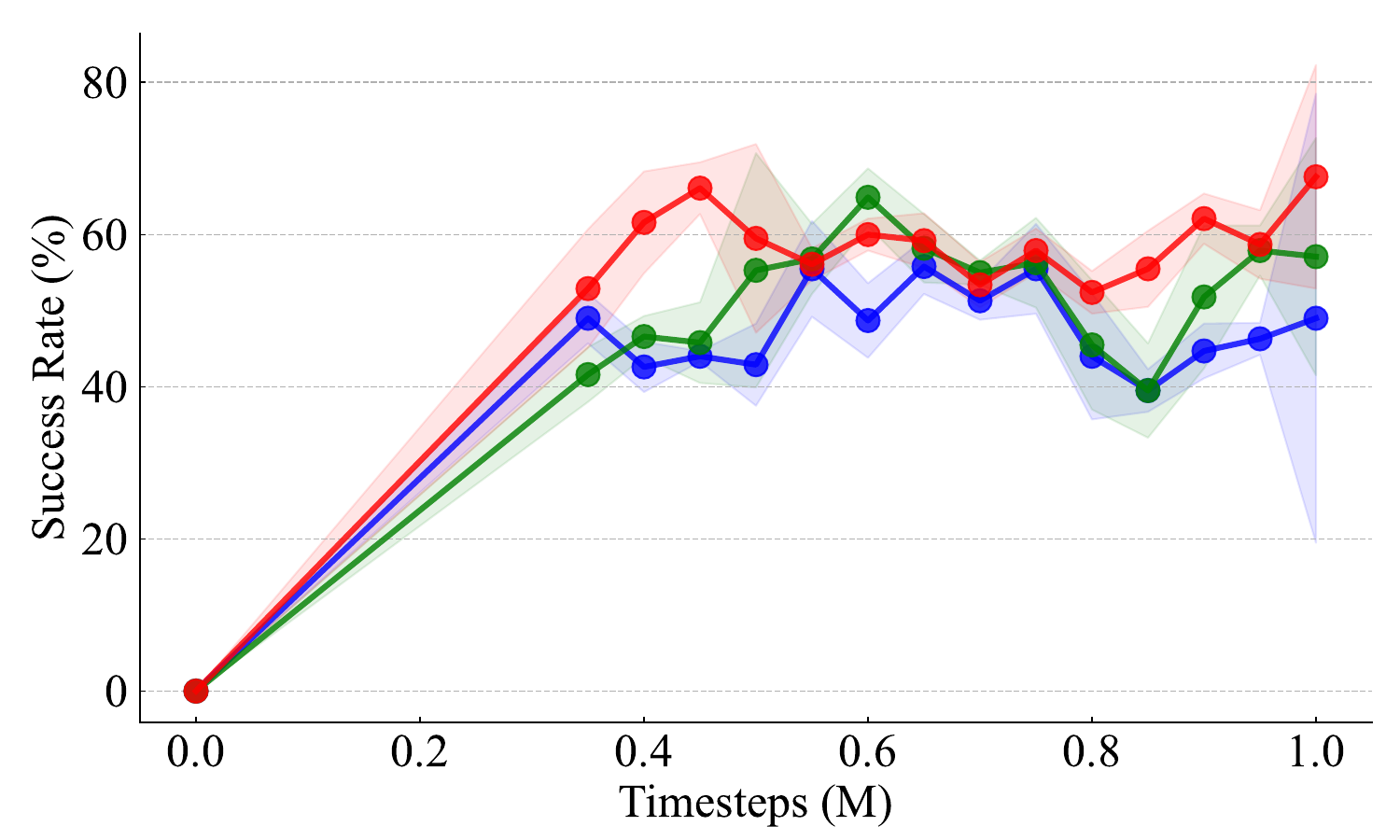}
        \label{sub:fig::efficiency:test}
    }
    \caption{Sample-efficiency of Prompt Ensemble-based Policy Learning (up to 1 million timesteps). These detailed evaluation graphs focused on the initial part of the training, are consistent with the experiment in Figure 4 of the manuscript.}
    \label{sub:fig:efficiency}
\end{figure*}

\subsection{Semantic Regularized Data Augmentation}
Table~\ref{sub:tab:semantic} shows the zero-shot performance for semantic regularized data augmentation in $\ourmodel$, where language data is additionally used. As shown, $\ourmodel$  with language data (w Semantic Reg.) consistently yields better performance over $\ourmodel$ without language data (w/o Semantic Reg.) across various noise scales ($\delta$).
As the deviation ($\delta$) of the Gaussian noise varies, it is observed that larger deviation does not necessarily lead to  performance improvement. This indicates that enhancing data augmentation diversity through higher noise deviation may not always be beneficial. However, when maintaining the semantics (w Semantic Reg.), the performance can improve with larger deviation.

\begin{table}[h]
\caption{Semantic Regularized Data Augmentation.}
\label{sub:tab:semantic}
\centering
\footnotesize
\adjustbox{max width=0.99\linewidth}{
\begin{tabular}{l ccc ccc}
\toprule
\multirow{2}{*}{$\delta$}
  & \multicolumn{3}{c}{w/o Semantic Reg.} & \multicolumn{3}{c}{w Semantic Reg.}\\
  \cmidrule(rl){2-4} \cmidrule(rl){5-7}
  & Source & Seen Target & Unseen Target & Source & Seen Target & Unseen Target\\
\midrule
0
& $90.8{\pm10.9}\%$
& $72.1{\pm6.5}\%$
& $80.0{\pm7.2}\%$

& $90.8{\pm10.9}\%$
& $78.0{\pm9.6}\%$
& $80.0{\pm7.2}\%$ \\

0.1
& $97.4{\pm3.8}\%$
& $84.5{\pm8.3}\%$
& $82.7{\pm9.4}\%$

& $\mathbf{100.0{\pm0.0}}\%$
& $\mathbf{86.0{\pm6.2}}\%$
& $\mathbf{82.1{\pm14.2}}\%$ \\

0.2
& $94.7{\pm0.0}\%$
& $82.1{\pm9.5}\%$
& $73.1{\pm16.1}\%$ 

& $\mathbf{94.8{\pm7.4}}\%$
& $\mathbf{84.2{\pm6.2}}\%$
& $\mathbf{75.1{\pm11.9}}\%$ \\

0.3
& $84.2{\pm3.7}\%$
& $77.4{\pm6.6}\%$
& $73.1{\pm10.9}\%$

& $\mathbf{96.1{\pm1.9}}\%$
& $\mathbf{86.1{\pm4.7}}\%$
& $\mathbf{80.1{\pm15.2}}\%$ \\

0.4
& $80.3{\pm16.2}\%$
& $75.8{\pm12.1}\%$
& $72.2{\pm17.7}\%$

& $\mathbf{86.9{\pm3.8}}\%$
& $\mathbf{80.5{\pm8.6}}\%$
& $\mathbf{76.0{\pm15.8}}\%$ \\

0.5
& $71.1{\pm9.6}\%$
& $64.5{\pm12.8}\%$
& $59.1{\pm14.2}\%$

& $\mathbf{73.7{\pm3.8}}\%$
& $\mathbf{68.4{\pm4.6}}\%$
& $\mathbf{66.7{\pm9.8}}\%$ \\
\bottomrule
\end{tabular}
}
\end{table}

\begin{table*}[p]
\caption{AI2THOR Expert Dataset}
\label{tab:AI2THOR:dataset}
\begin{center}
\begin{small}
\adjustbox{max width=\textwidth}{
\begin{tabular}{lccc cccc cccc cc}
\toprule
 \multirow{2}{*}{ID} & \multirow{2}{*}{Observation} & \multicolumn{4}{c}{Environmental Difference} & \multicolumn{4}{c}{Physical Property} & \multicolumn{2}{c}{Expert Episode}\\
 \cmidrule(rl){3-6} \cmidrule(rl){7-10} \cmidrule(rl){11-12}
 &  & Brightness & Contrast & Saturation & Hue & Camera & Step Size & Rotation Degree & Look Degree & SPL & Length\\
\midrule
 \multirow{5}{*}{1} 
 & \multirow{5}{*}{\includegraphics[scale=0.15]{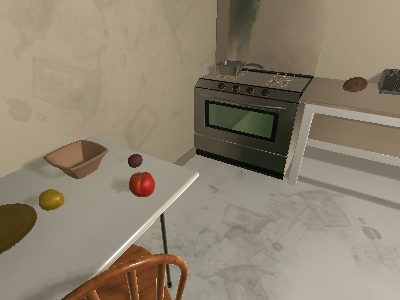}} 
 & \multirow{5}{*}{1.0} & \multirow{5}{*}{1.0} & \multirow{5}{*}{1.0} & \multirow{5}{*}{0.0}
 & \multirow{5}{*}{63.5} & \multirow{5}{*}{0.25} & \multirow{5}{*}{30.0\textdegree} & \multirow{5}{*}{30.0\textdegree}
 & \multirow{5}{*}{0.90} & \multirow{5}{*}{9.58} \\ 
 \\
 \\
 \\
 \\
 \multirow{5}{*}{2} & \multirow{5}{*}{\includegraphics[scale=0.15]{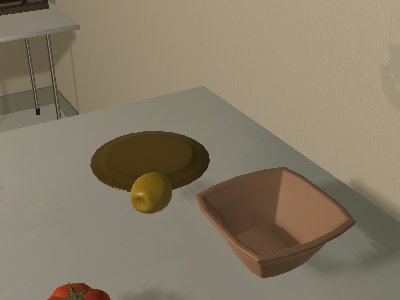}} 
 & \multirow{5}{*}{1.0} & \multirow{5}{*}{1.0} & \multirow{5}{*}{1.0} & \multirow{5}{*}{0.0}
 & \multirow{5}{*}{29.7} & \multirow{5}{*}{0.25} & \multirow{5}{*}{30.0\textdegree} & \multirow{5}{*}{30.0\textdegree}
 & \multirow{5}{*}{0.92} & \multirow{5}{*}{23.30} \\ 
 \\
 \\
 \\
 \\
 \multirow{5}{*}{3} & \multirow{5}{*}{\includegraphics[scale=0.15]{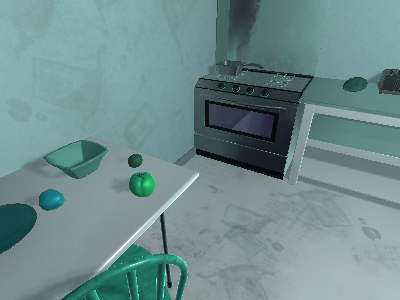}} 
 & \multirow{5}{*}{1.0} & \multirow{5}{*}{1.0} & \multirow{5}{*}{1.0} & \multirow{5}{*}{0.4}
 & \multirow{5}{*}{63.5} & \multirow{5}{*}{0.25} & \multirow{5}{*}{30.0\textdegree} & \multirow{5}{*}{30.0\textdegree}
 & \multirow{5}{*}{0.90} & \multirow{5}{*}{9.50} \\ 
 \\
 \\
 \\
 \\

 \multirow{5}{*}{4} & \multirow{5}{*}{\includegraphics[scale=0.15]{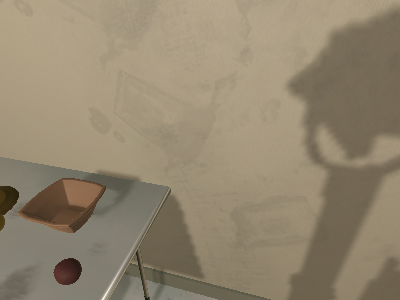}}
 & \multirow{5}{*}{1.0} & \multirow{5}{*}{1.0} & \multirow{5}{*}{1.0} & \multirow{5}{*}{0.0}
 & \multirow{5}{*}{63.5} & \multirow{5}{*}{0.25} & \multirow{5}{*}{5.0\textdegree} & \multirow{5}{*}{30.0\textdegree}
 & \multirow{5}{*}{0.91} & \multirow{5}{*}{24.90} \\ 
 \\
 \\
 \\
 \\
 \multirow{5}{*}{5} & \multirow{5}{*}{\includegraphics[scale=0.15]{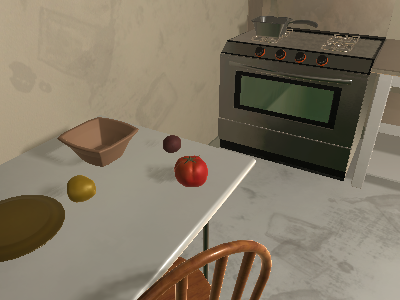}} 
 & \multirow{5}{*}{1.0} & \multirow{5}{*}{1.0} & \multirow{5}{*}{1.0} & \multirow{5}{*}{0.0}
 & \multirow{5}{*}{46.0} & \multirow{5}{*}{0.25} & \multirow{5}{*}{30.0\textdegree} & \multirow{5}{*}{30.0\textdegree}
 & \multirow{5}{*}{0.88} & \multirow{5}{*}{35.80} \\ 
 \\
 \\
 \\
 \\
 \multirow{5}{*}{6} & \multirow{5}{*}{\includegraphics[scale=0.15]{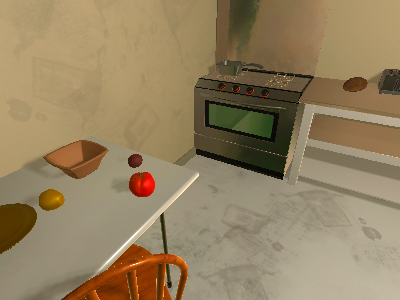}} 
 & \multirow{5}{*}{1.0} & \multirow{5}{*}{1.0} & \multirow{5}{*}{1.7} & \multirow{5}{*}{0.0}
 & \multirow{5}{*}{63.5} & \multirow{5}{*}{0.25} & \multirow{5}{*}{30.0\textdegree} & \multirow{5}{*}{30.0\textdegree}
 & \multirow{5}{*}{0.90} & \multirow{5}{*}{9.58} \\ 
 \\
 \\
 \\
 \\
 \multirow{5}{*}{7} & \multirow{5}{*}{\includegraphics[scale=0.15]{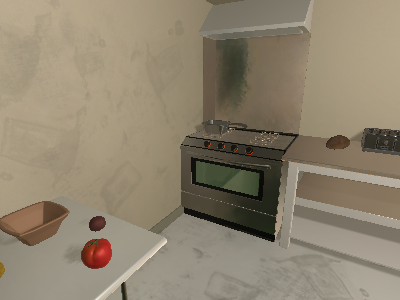}} 
 & \multirow{5}{*}{1.0} & \multirow{5}{*}{1.0} & \multirow{5}{*}{1.0} & \multirow{5}{*}{0.0}
 & \multirow{5}{*}{63.5} & \multirow{5}{*}{0.01} & \multirow{5}{*}{30.0\textdegree} & \multirow{5}{*}{30.0\textdegree}
 & \multirow{5}{*}{0.97} & \multirow{5}{*}{110.0} \\ 
 \\
 \\
 \\
 \\
 \multirow{5}{*}{8} & \multirow{5}{*}{\includegraphics[scale=0.15]{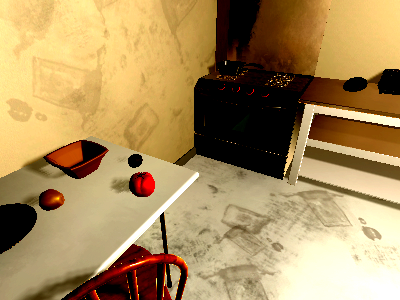}}
 & \multirow{5}{*}{1.0} & \multirow{5}{*}{3.3} & \multirow{5}{*}{1.0} & \multirow{5}{*}{0.0}
 & \multirow{5}{*}{63.5} & \multirow{5}{*}{0.25} & \multirow{5}{*}{30.0\textdegree} & \multirow{5}{*}{30.0\textdegree}
 & \multirow{5}{*}{0.90} & \multirow{5}{*}{9.58} \\ 
 \\
 \\
 \\
 \\
 \multirow{5}{*}{9} & \multirow{5}{*}{\includegraphics[scale=0.15]{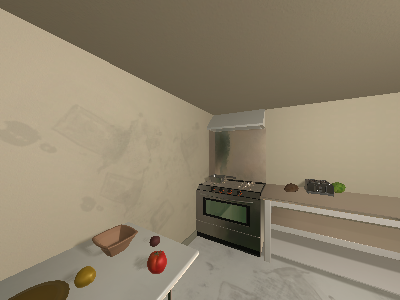}}
 & \multirow{5}{*}{1.0} & \multirow{5}{*}{1.0} & \multirow{5}{*}{1.0} & \multirow{5}{*}{0.0}
 & \multirow{5}{*}{92.9} & \multirow{5}{*}{0.25} & \multirow{5}{*}{30.0\textdegree} & \multirow{5}{*}{30.0\textdegree}
 & \multirow{5}{*}{0.84} & \multirow{5}{*}{8.82} \\ 
 \\
 \\
 \\
 \\
\multirow{5}{*}{10} & \multirow{5}{*}{\includegraphics[scale=0.15]{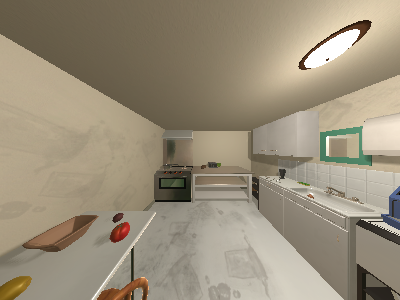}} 
 & \multirow{5}{*}{1.0} & \multirow{5}{*}{1.0} & \multirow{5}{*}{1.0} & \multirow{5}{*}{0.0}
 & \multirow{5}{*}{127.0} & \multirow{5}{*}{0.25} & \multirow{5}{*}{30.0\textdegree} & \multirow{5}{*}{30.0\textdegree}
 & \multirow{5}{*}{0.82} & \multirow{5}{*}{8.58} \\ 
\\
\\
\\
\\
\bottomrule
\end{tabular}
}
\end{small}
\end{center}
\end{table*}

\begin{table*}[p]
\caption{Egocentric-Metaworld Expert Dataset}
\label{tab:Metaworld:dataset}
\begin{center}
\begin{small}
\adjustbox{max width=\textwidth}{
\begin{tabular}{lc c cccc ccc cc}
\toprule
 \multirow{2}{*}{ID} & \multirow{2}{*}{Observation} & \multicolumn{4}{c}{Environmental Difference} & \multicolumn{3}{c}{Physical Property} & \multicolumn{2}{c}{Expert Episode}\\
\cmidrule(rl){3-6} \cmidrule(rl){7-9} \cmidrule(rl){10-11}
 &  & Brightness & Contrast & Saturation & Hue & Camera Position & Wind & Gravity & Rewards & Length\\
\midrule

\multirow{6}{*}{1} & \multirow{6}{*}{\includegraphics[scale=0.25]{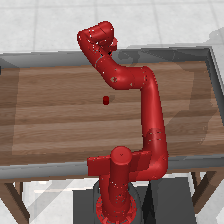}} 
& \multirow{6}{*}{1.0} & \multirow{6}{*}{1.0} & \multirow{6}{*}{1.0} & \multirow{6}{*}{0.0} 
& \multirow{6}{*}{1.0} & \multirow{6}{*}{0.0} & \multirow{6}{*}{0.0}
& \multirow{6}{*}{315.48} & \multirow{6}{*}{48.00}\\
  \\
  \\
  \\
  \\
  \\
 \multirow{6}{*}{2} & \multirow{6}{*}{\includegraphics[scale=0.25]{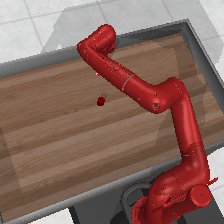}} 
& \multirow{6}{*}{1.0} & \multirow{6}{*}{1.0} & \multirow{6}{*}{1.0} & \multirow{6}{*}{0.0} 
& \multirow{6}{*}{2.0} & \multirow{6}{*}{0.0} & \multirow{6}{*}{0.0}
& \multirow{6}{*}{315.48} & \multirow{6}{*}{48.00}\\
 \\
 \\
 \\
 \\
 \\
 \multirow{6}{*}{3} & \multirow{6}{*}{\includegraphics[scale=0.25]{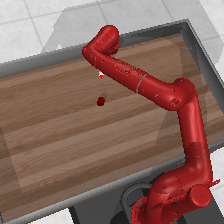}} 
& \multirow{6}{*}{1.0} & \multirow{6}{*}{1.0} & \multirow{6}{*}{1.0} & \multirow{6}{*}{0.0} 
& \multirow{6}{*}{2.0} & \multirow{6}{*}{0.0} & \multirow{6}{*}{9.0}
& \multirow{6}{*}{199.14} & \multirow{6}{*}{35.00}\\
 \\
 \\
 \\
 \\
 \\
 \multirow{6}{*}{4} & \multirow{6}{*}{\includegraphics[scale=0.25]{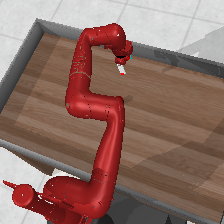}} 
& \multirow{6}{*}{1.0} & \multirow{6}{*}{1.0} & \multirow{6}{*}{1.0} & \multirow{6}{*}{0.0} 
& \multirow{6}{*}{3.0} & \multirow{6}{*}{0.0} & \multirow{6}{*}{1.0}
& \multirow{6}{*}{315.48} & \multirow{6}{*}{48.00}\\
 \\
 \\
 \\
 \\
 \\
 \multirow{6}{*}{5} & \multirow{6}{*}{\includegraphics[scale=0.25]{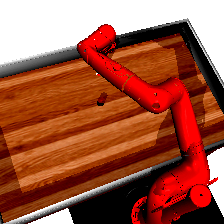}} 
& \multirow{6}{*}{1.0} & \multirow{6}{*}{3.3} & \multirow{6}{*}{1.0} & \multirow{6}{*}{0.0} 
& \multirow{6}{*}{2.0} & \multirow{6}{*}{0.0} & \multirow{6}{*}{1.0}
& \multirow{6}{*}{315.48} & \multirow{6}{*}{48.00}\\
 \\
 \\
 \\
 \\
 \\
 \multirow{6}{*}{6} & \multirow{6}{*}{\includegraphics[scale=0.25]{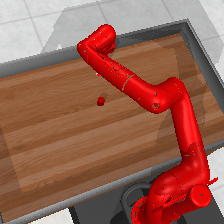}} 
& \multirow{6}{*}{1.0} & \multirow{6}{*}{1.0} & \multirow{6}{*}{1.7} & \multirow{6}{*}{0.0} 
& \multirow{6}{*}{2.0} & \multirow{6}{*}{0.0} & \multirow{6}{*}{1.0}
& \multirow{6}{*}{315.48} & \multirow{6}{*}{48.00}\\
 \\
 \\
 \\
 \\
 \\
 \multirow{6}{*}{7} & \multirow{6}{*}{\includegraphics[scale=0.25]{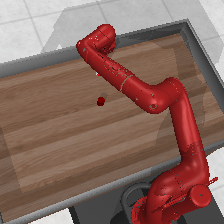}} 
& \multirow{6}{*}{1.0} & \multirow{6}{*}{1.0} & \multirow{6}{*}{1.0} & \multirow{6}{*}{0.0} 
& \multirow{6}{*}{2.0} & \multirow{6}{*}{8.0} & \multirow{6}{*}{1.0}
& \multirow{6}{*}{387.49} & \multirow{6}{*}{56.00}\\
 \\
 \\
 \\
 \\
 \\
 \multirow{6}{*}{8} & \multirow{6}{*}{\includegraphics[scale=0.25]{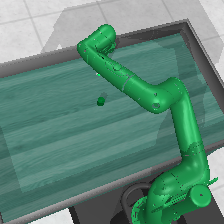}} 
& \multirow{6}{*}{1.0} & \multirow{6}{*}{1.0} & \multirow{6}{*}{1.0} & \multirow{6}{*}{0.4} 
& \multirow{6}{*}{2.0} & \multirow{6}{*}{0.0} & \multirow{6}{*}{1.0}
& \multirow{6}{*}{315.48} & \multirow{6}{*}{48.00}\\
 \\
 \\
 \\
 \\
 \\
 \multirow{6}{*}{9} & \multirow{6}{*}{\includegraphics[scale=0.25]{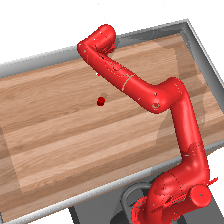}} 
& \multirow{6}{*}{1.5} & \multirow{6}{*}{1.0} & \multirow{6}{*}{1.0} & \multirow{6}{*}{0.0} 
& \multirow{6}{*}{2.0} & \multirow{6}{*}{0.0} & \multirow{6}{*}{1.0}
& \multirow{6}{*}{315.48} & \multirow{6}{*}{48.00}\\
 \\
 \\
 \\
 \\
 \\
  \multirow{6}{*}{10} & \multirow{6}{*}{\includegraphics[scale=0.25]{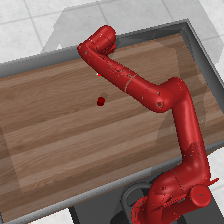}} 
& \multirow{6}{*}{1.0} & \multirow{6}{*}{1.0} & \multirow{6}{*}{1.0} & \multirow{6}{*}{0.0} 
& \multirow{6}{*}{2.0} & \multirow{6}{*}{0.0} & \multirow{6}{*}{1.0}
& \multirow{6}{*}{252.86} & \multirow{6}{*}{41.00}\\
 \\
 \\
 \\
 \\
 \\
 \bottomrule
\end{tabular}
}
\end{small}
\end{center}
\end{table*}

\begin{table*}[p]
\caption{CARLA Expert Dataset}
\label{tab:CARA:dataset}
\begin{center}
\begin{small}
\adjustbox{max width=\textwidth}{
\begin{tabular}{lc c cc ccc cc}
\toprule
 \multirow{2}{*}{ID} & \multirow{2}{*}{Observation} & \multicolumn{2}{c}{Environmental Difference} & \multicolumn{3}{c}{Physical Property} & \multicolumn{2}{c}{Expert Episode}\\
\cmidrule(rl){3-4} \cmidrule(rl){5-7} \cmidrule(rl){8-9}
 &  & Weather & Daytime & Control Sensitivity & Camera Position & Field of View & Rewards & Length\\
\midrule
\multirow{6}{*}{1} & \multirow{6}{*}{\includegraphics[scale=0.25]{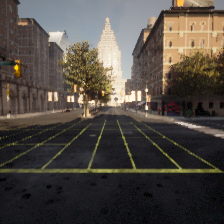}} 
& \multirow{6}{*}{Clear} & \multirow{6}{*}{Sunset} 
& \multirow{6}{*}{(100\%, 100\%, 100\%)} & \multirow{6}{*}{high} & \multirow{6}{*}{75 \textdegree}
& \multirow{6}{*}{2691.8} & \multirow{6}{*}{758}\\
  \\
  \\
  \\
  \\
  \\
 \multirow{6}{*}{2} & \multirow{6}{*}{\includegraphics[scale=0.25]{figures/carla/df2.png}} 
 & \multirow{6}{*}{Clear} & \multirow{6}{*}{Noon} 
 & \multirow{6}{*}{(100\%, 85\%, 100\%)} & \multirow{6}{*}{low} & \multirow{6}{*}{60 \textdegree} 
 & \multirow{6}{*}{2713.2} & \multirow{6}{*}{749} \\ 
 \\
 \\
 \\
 \\
 \\
 \multirow{6}{*}{3} & \multirow{6}{*}{\includegraphics[scale=0.25]{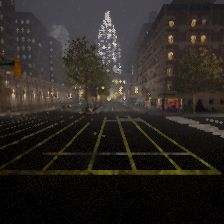}} 
 & \multirow{6}{*}{Cloudy} & \multirow{6}{*}{Night} 
 & \multirow{6}{*}{(85\%, 100\%, 85\%)} & \multirow{6}{*}{low} & \multirow{6}{*}{90 \textdegree}
 & \multirow{6}{*}{2747.9} & \multirow{6}{*}{733} \\
 \\
 \\
 \\
 \\
 \\
 \multirow{6}{*}{4} & \multirow{6}{*}{\includegraphics[scale=0.25]{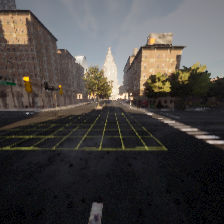}} 
 & \multirow{6}{*}{Clear} & \multirow{6}{*}{Sunset} 
 & \multirow{6}{*}{(100\%, 100\%,100\%)} & \multirow{6}{*}{high} & \multirow{6}{*}{110 \textdegree} 
 & \multirow{6}{*}{2742.8} & \multirow{6}{*}{733} \\
 \\
 \\
 \\
 \\
 \\
 \multirow{6}{*}{5} & \multirow{6}{*}{\includegraphics[scale=0.25]{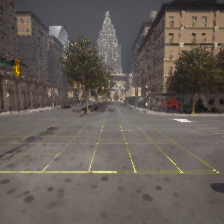}} 
 & \multirow{6}{*}{Cloudy} & \multirow{6}{*}{Noon} 
 & \multirow{6}{*}{(100\%, 85\%, 100\%)} & \multirow{6}{*}{low} & \multirow{6}{*}{60 \textdegree} 
& \multirow{6}{*}{2746.6} & \multirow{6}{*}{736}  \\
 \\
 \\
 \\
 \\
 \\
 \multirow{6}{*}{6} & \multirow{6}{*}{\includegraphics[scale=0.25]{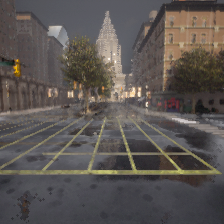}} 
 & \multirow{6}{*}{MidRainy} & \multirow{6}{*}{Sunset} 
 & \multirow{6}{*}{(70\%, 70\%, 85\%)} & \multirow{6}{*}{low} & \multirow{6}{*}{60 \textdegree} 
 & \multirow{6}{*}{2736.9} & \multirow{6}{*}{734} \\ 
 \\
 \\
 \\
 \\
 \\
 \multirow{6}{*}{7} & \multirow{6}{*}{\includegraphics[scale=0.25]{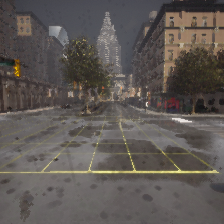}} 
 & \multirow{6}{*}{MidRainy} & \multirow{6}{*}{Noon} 
 & \multirow{6}{*}{(100\%, 85\%, 70\%)} & \multirow{6}{*}{low} & \multirow{6}{*}{60 \textdegree} 
 & \multirow{6}{*}{2735.2} & \multirow{6}{*}{736}  \\
 \\
 \\
 \\
 \\
 \\
 \multirow{6}{*}{8} & \multirow{6}{*}{\includegraphics[scale=0.25]{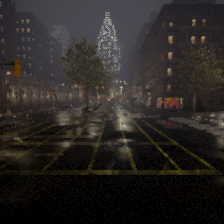}} 
 & \multirow{6}{*}{SoftRainy} & \multirow{6}{*}{Nigjht} 
 & \multirow{6}{*}{(70\%, 70\%, 70\%)} & \multirow{6}{*}{low} & \multirow{6}{*}{60 \textdegree} 
 & \multirow{6}{*}{2716.8} & \multirow{6}{*}{746}  \\
 \\
 \\
 \\
 \\
 \\
 \multirow{6}{*}{9} & \multirow{6}{*}{\includegraphics[scale=0.25]{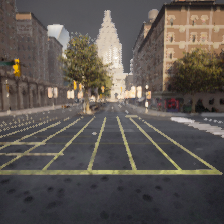}} 
 & \multirow{6}{*}{Cloudy} & \multirow{6}{*}{Sunset} 
 & \multirow{6}{*}{(70\%, 70\%, 100\%)} & \multirow{6}{*}{low} & \multirow{6}{*}{60 \textdegree} 
 & \multirow{6}{*}{2739.8} & \multirow{6}{*}{731}  \\
 \\
 \\
 \\
 \\
 \\
  \multirow{6}{*}{10} & \multirow{6}{*}{\includegraphics[scale=0.25]{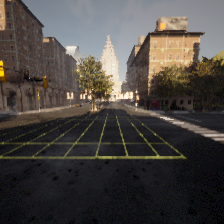}} 
 & \multirow{6}{*}{Clear} & \multirow{6}{*}{Sunset} 
 & \multirow{6}{*}{(100\%, 100\%, 100\%)} & \multirow{6}{*}{high} & \multirow{6}{*}{95 \textdegree} 
 & \multirow{6}{*}{2725.7} & \multirow{6}{*}{738}  \\
 \\
 \\
 \\
 \\
 \\
 \bottomrule
\end{tabular}
}
\end{small}
\end{center}
\end{table*}

\printbibliography
\end{document}